\documentclass[10pt,journal,compsoc]{IEEEtran}
\usepackage{amsmath,amsfonts}
\usepackage{algorithmic}
\usepackage{algorithm}
\usepackage{array}
\usepackage[caption=false,font=normalsize,labelfont=sf,textfont=sf]{subfig}
\usepackage{textcomp}
\usepackage{stfloats}
\usepackage{url}
\usepackage{verbatim}
\usepackage{booktabs}
\usepackage{caption}
\usepackage{multirow}
\usepackage{overpic}
\usepackage{color}
\usepackage{graphicx}
\usepackage{cite}
\begin{document}

\title{ImmersiveNeRF:\\Hybrid Radiance Fields for Unbounded Immersive Light Field Reconstruction}

\author{Xiaohang~Yu, Haoxiang ~Wang, Yuqi~Han, Lei~Yang, Tao~Yu,  and Qionghai~Dai 
\IEEEcompsocitemizethanks{
\IEEEcompsocthanksitem X. Y, H. W, Y. H, T. Y, and Q. D are
with Tsinghua University, Beijing 100084, China. L. Y is with SenseTime Technology \protect\\
E-mail: 
yuxh21@mails.tsinghua.edu.cn, whx22@mails.tsinghua.edu.cn,\\
yqhan@mail.tsinghua.edu.cn,yanglei@sensetime.com,\\
ytrock@126.com, daiqionghai@tsinghua.edu.cn.
}
}

\markboth{Journal of \LaTeX\ Class Files,~Vol.~14, No.~8, August~2021}%
{Shell \MakeLowercase{\textit{et al.}}: A Sample Article Using IEEEtran.cls for IEEE Journals}


\IEEEtitleabstractindextext{
\begin{abstract}
This paper proposes a hybrid radiance field representation for unbounded immersive light field reconstruction which supports high-quality rendering and aggressive view extrapolation. The key idea is to first formally separate the foreground and the background and then adaptively balance learning of them during the training process. To fulfill this goal, we represent the foreground and background as two separate radiance fields with two different spatial mapping strategies. We further propose an adaptive sampling strategy and a segmentation regularizer for more clear segmentation and robust convergence. Finally, we contribute a novel immersive light field dataset, named THUImmersive, with the potential to achieve much larger space 6DoF immersive rendering effects compared with existing datasets, by capturing multiple neighboring viewpoints for the same scene, to stimulate the research and AR/VR applications in the immersive light field domain. Extensive experiments demonstrate the strong performance of our method for unbounded immersive light field reconstruction. 
\end{abstract}

}
\maketitle
\IEEEdisplaynontitleabstractindextext
\IEEEpeerreviewmaketitle
\section{Introduction}
Immersive light field acquisition and reconstruction is one of the fundamental technologies supporting photo-realistic AR and VR experience. Immersive light fields can be captured by uniformly distributed cameras on a sphere in an inside-looking-out configuration, collecting both foreground and background information of the scene and supporting a certain range, usually within a one-meter diameter hemisphere, to 6DOF novel view rendering. Compared to traditional 360 panorama cameras, the immersive light fields can produce physically correct parallax change effects, which significantly enhance the sense of immersion. 

Google first built the immersive light field system \cite{broxton2020immersive} in 2020. \cite{broxton2020immersive} proposed MultiSphereImage (MSI) for representing and optimizing the immersive light field, and achieved a plausible immersive viewing effect in VR. However, the MSI-based discrete representation leads to reconstruction algorithms that are sensitive to the number of multi-sphere images used, which limits the rendering resolution and results in the presence of artifacts such as ghosts near the scene boundary.

\begin{figure*}[htbp]
\centering
\begin{tabular}{cccc}
\begin{tabular}{c}

    \includegraphics[height=0.15\linewidth]{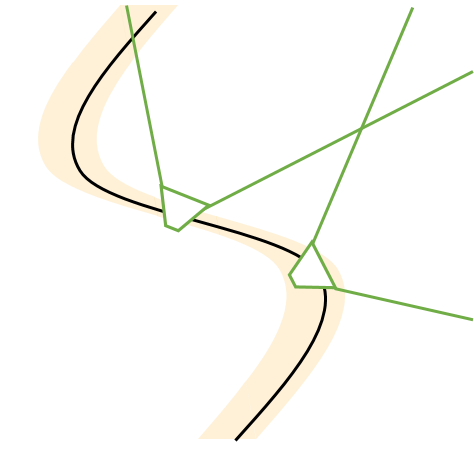} 
    \\(a) 
\end{tabular}&
\begin{tabular}{c}
    \includegraphics[height=0.15\linewidth]{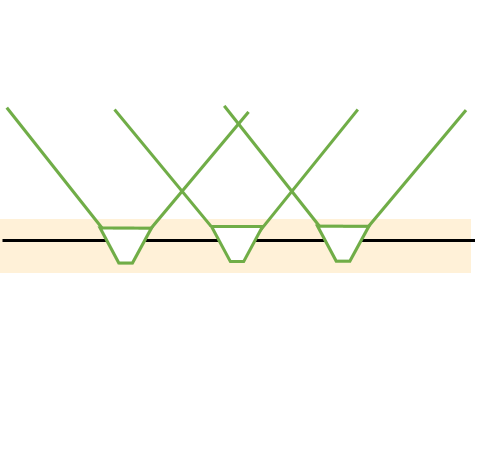} 
    \\(b)
\end{tabular}&
\begin{tabular}{c}
    \includegraphics[height=0.15\linewidth]{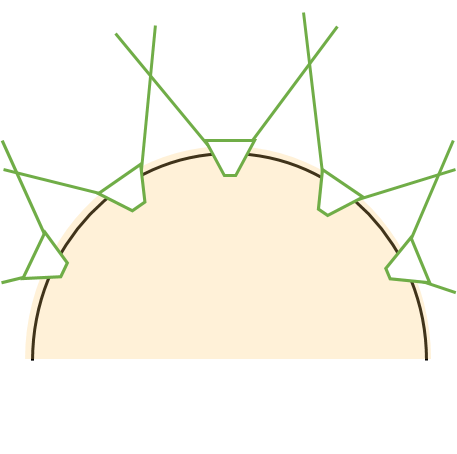} 
    \\(c)
\end{tabular} &
\begin{tabular}{c}
    \includegraphics[height=0.15\linewidth]{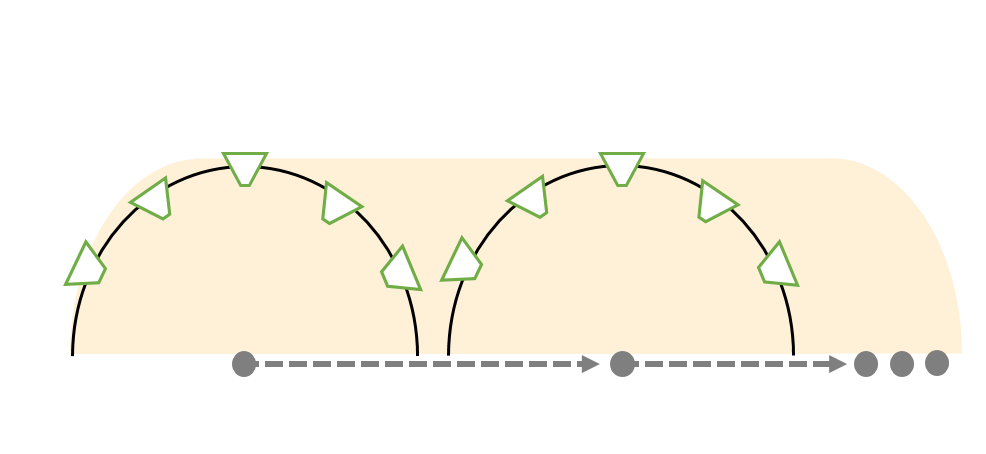} 
    \\(d)
\end{tabular}
\end{tabular}
\caption{Different camera setup and the corresponding valid NVS volume from top view. 
From left to right: (a) single camera trajectory, (b) planar light field, (c) domed light field, (d) domed light field trajectory.}
\label{figure:topview}
\end{figure*}
In order to overcome the limitations of discrete representations, NeRF \cite{mildenhall2021nerf} proposes a continuous implicit function for a unified representation of the density field and color field of the scene to achieve realistic novel viewpoint synthesis performance. 
However, for large unbounded immersive light fields, the vanilla NeRF still suffers from unstable optimization and low rendering quality. The reason is that the vanilla NeRF does not consider any spatial mapping strategy for unbounded scenes, which makes it difficult to learn a consistent multi-viewpoint representation for regions in the background where the parallax is close to $0$. This will further affect the learning of foreground regions, leading to problems such as decreased effective rendering resolution. 
Although both NeRF++ \cite{zhang2020nerf++} and Mip-NeRF-360  \cite{barron2022mip} improve the performance of the vanilla NeRF for unbounded scene reconstruction, neither performs well enough for the immersive light field, a scenario with a relatively large angle-baseline that contains both foreground and unbounded background. We attribute this phenomenon to the fact that the parallax distribution becomes more inhomogeneous in the presence of sparse inputs, making it more difficult to train with a uniform sampling strategy even with unbounded spatial mapping.

To resolve the limitations above, we propose ImmersiveNeRF, a neural representation specifically designed for unbounded immersive light field reconstruction. 
Our key idea is that for unbounded immersive light fields, the foreground and background should be explicitly represented and learned in different fields. Specifically, we argue that the foreground is better suited for learning in Euclidean space, while the background, where parallax is small, is better suited for learning in the mapped spherical coordinate space. 
To this end, we propose a hybrid radiance field scene representation, which explicitly decomposes the unbounded scene into two radiance fields in two different coordinates. We further propose an adaptive sampling strategy to guarantee that the learning of the background with respect to the foreground is more uniform and smooth. Last but not least, we propose a new regularizer to ensure that the learning process of the hybrid radiance fields can automatically split the foreground and background, clearly eliminate the impact of the background on the foreground, and further improve both the rendering and depth reconstruction quality of the foreground radiance field.
Extensive experiments demonstrate the effectiveness of ImmersiveNeRF for high-quality unbounded immersive light field novel view synthesis even for aggressive extrapolation viewpoints. 
Finally, to advance the research and application of immersive light field reconstruction, we built an acquisition system and acquired a new immersive light field dataset, named THUImmersive. Different from the existing Immersive Light Field dataset \cite{broxton2020immersive} that only acquires one viewpoint in a scene, we collected multiple neighboring viewpoints in the same scene, each of which is an immersive light field, covering a much larger 3D space, thus encouraging subsequent algorithms to use the immersive light field data from multiple viewpoints simultaneously for reconstruction to support much larger range (not a 6DOF with a fixed position where only the head can be moved, but a 6DOF rendering that allows people to move and explore) for an immersive and unbounded scene viewing experience. 

For simplicity, we conduct our analysis in 2D from the top view in Fig. \ref{figure:topview}. 
The figure shows camera viewing frusta (green wedges) for cameras on different trajectories (black solid line), and the corresponding valid NVS (novel view synthesis) volume. We define valid NVS volume as the region where the rendering effects do not deviate far from that of the training views. Subfigure (a) indicates single camera takes pictures at certain points along a trajectory. Dataset Tanks and Temples \cite{tanks} and dataset used in \cite{rosinol2022nerfslam} follow the capture setup in Subfigure (a).
Subfigure (b) and (c) show planar \cite{mildenhall2019local} and domed \cite{broxton2020immersive} multi-camera array, respectively. Subfigure (d) illustrates that regarding the domed multi-camera array as a single camera with about $220^\circ$ FOV (green semicircle), we take pictures at certain points along a trajectory.   
To show the scope of novel view synthesis with high quality, we label the valid NVS volume in orange shadow.
Taking only the camera setup into account, the interpolation volume is only decided by the curvature of the trajectory and FOV for each camera. As it is shown in Fig. \ref{figure:topview}, a domed light field on a trajectory can efficiently derive the largest and most flexible valid NVS volume than others.

Our contributions can be summarized as:
\begin{itemize}
\item We propose a hybrid radiance field representation for unbounded immersive light field reconstruction which can decompose the foreground and background regions robustly and automatically during the training process. 

\item We propose an adaptive sampling strategy to balance the learning efficiency of the foreground and the background, thus ensuring a more robust convergence performance. Moreover, a simple but effective regularizer was proposed to enhance the segmentation of the foreground to eliminate the negative effects caused by background colors. 

\item A challenging dataset for the unbounded immersive light field is captured to encourage extra-large scale 6DOF immersive rendering performance, which we believe will stimulate related research.  
\end{itemize}

\begin{figure*}[htbp]
\centering
\includegraphics[width=0.95\textwidth]{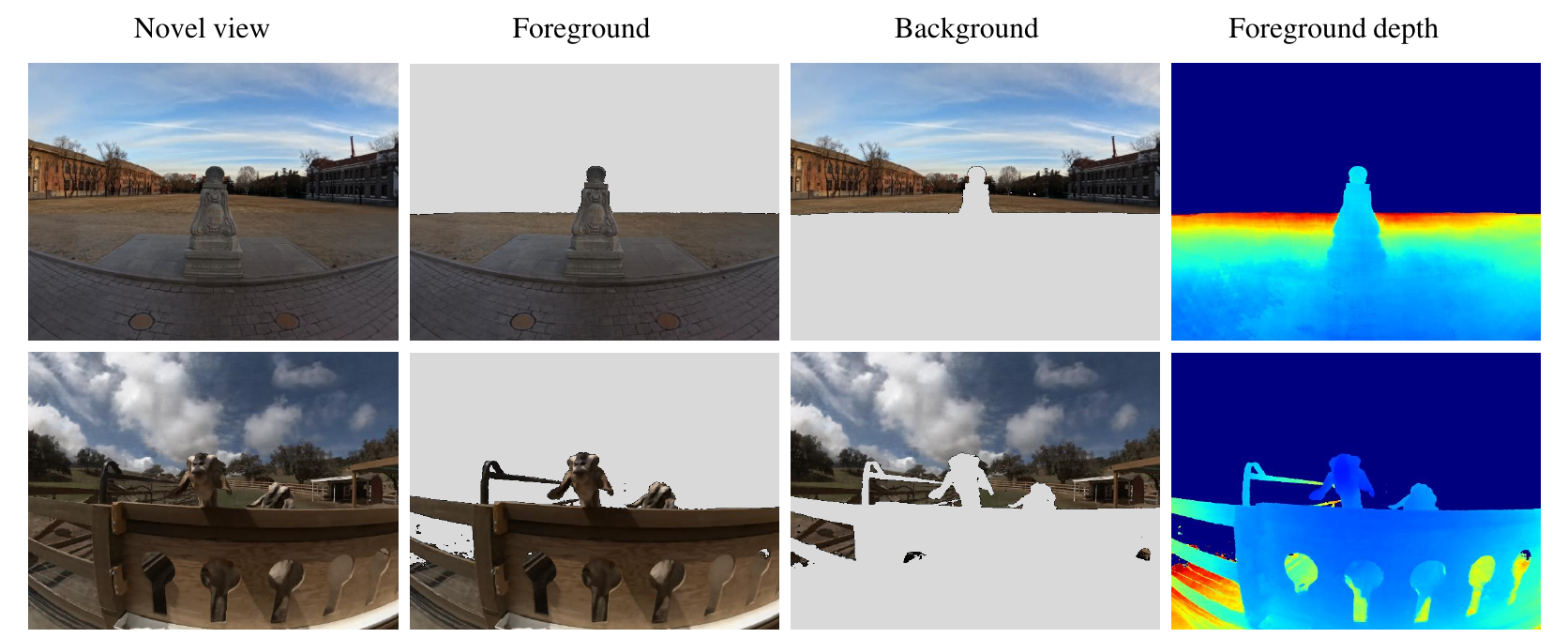}
\caption{ We propose ImmersiveNeRF, a hybrid representation for novel view rendering in outdoor unbounded scenes. ImmersiveNeRF takes a set of inside-looking-out multi-view images as input and learns separate foreground and background fields according to the disparity from input images for high-quality and large space (with aggressive extrapolation) 6DoF novel view synthesis that supports AR/VR immersive experience.}
\label{figure:teaser}
\end{figure*}

\section{Related Works}

Our work is related to several research domains, such as large-scale 3D reconstruction, omnidirectional novel view synthesis, and efficient sampling for volume rendering.
\subsection{Large scale 3D reconstruction}
Large unbounded scene reconstruction has always been one of the most challenging problems in computer vision. In the early times, the large-scale view synthesis was based on a large number of images warping and interpolating \cite{aliaga2002sea}. Some previous works aim at the architecture reconstruction with some geometry priors \cite{buehler2001unstructured, debevec1996modeling}. Some GPU-friendly data structures are also proposed for the large scene reconstruction \cite{chen2013scalable, niessner2013real}. 
Later some fusion or warping methods based on learning feature extraction for the images are proposed \cite{vineet2015incremental,hedman2018deep}. More viewpoints interpolation methods are also proposed based on the 2D deep learning models \cite{wang2021ibrnet,meshry2019neural}. Neural Radiance Fields (NeRF) \cite{mildenhall2021nerf} achieves higher reconstruction quality with a model-based volumetric rendering framework. MLP embedded in NeRF can well integrate efficient information in the sparse 3D space. The method has been soon extended to large scene reconstruction. NeRF-in-the wild is the pioneer of the NeRF-based outdoor large scenes rendering \cite{martin2021nerf}. Some followers like Block-NeRF \cite{tancik2022block}, Mega-Nerf \cite{turki2022mega}, Urban Radiance Field (URF) \cite{rematas2022urban}, CityNeRF \cite{xiangli2021citynerf} are also showing the high potential for NeRF in the large scenes reconstruction. Beyond the large size of the scene, some other frameworks are also proposed for the visual performance and geometry accuracy of the rendering results. Some typical ones are NeRF++ \cite{zhang2020nerf++} and Mip-NeRF-360 \cite{barron2022mip}. The former conducts a foreground and background decomposition with the aid of the double NeRFs and can divide the scenes with the disparity to achieve better visual effects. The latter works on the reparametrization of the representation of the scene and proposes a new sampling strategy for efficient NeRF training. Some techniques in the work are general for reducing some artifacts in the unbounded NeRF reconstruction. Due to the large scale of the scenes, some artifacts will also be noticeable in large-scale novel view extrapolation.
Some recent literature also refers to this problem. They often design learnable primitives for efficient NeRF training and fast rendering \cite{attal2023hyperreel,chen2022mobilenerf,reiser2023merf}. The design structure will provide high representation efficiency embedded with structural priors, benefitting large-scale scenes. However, these methods often care about the compactness of the model itself but do not divide the foreground and background. This will often make it difficult to get both parts well. Often good background priors can not be generalized to the foreground and it will lead to some failure cases for these approaches.

\subsection{Omni-directional view synthesis}
Panoramic view synthesis has also been widely applied in VR/AR to achieve 6 DOF rendering. One of the early 6 DOF panorama renderings with motion parallax was proposed by Serrano et al.\cite{8661657}. They predict the depth map and mesh representation of the scene. 
MODS\cite{attal2020matryodshka}, SOMSI\cite{habtegebrial2022somsi}, and Casual 6-DoF \cite{9779957} showed the effectiveness of multi-sphere images for ${360^\circ}$ view synthesis. And Omni-NeRF\cite{gu2022omni} established omni-directional radiance field using spherical images. However, these methods take panorama shot input at dense multiple viewpoints, and the overlap ratio between input images is almost 100\%. \cite{habtegebrial2022somsi} generates its dataset in the manner of a forward-looking light field with 25 sub-aperture panoramas. Therefore, the rendering quality and the movable range of viewpoint are limited. In this paper, We capture the scene in a relatively large baseline with a limited field of view. Our proposed method aims to cope with more sparse image input to provide large space 6DoF free-viewpoint rendering with high rendering quality. 

\subsection{Efficient sampling for volume rendering}
Sampling Strategy is always a problem in volume rendering. Some traditional methods are proposed to improve the efficiency of sampling, like inverse transform sampling \cite{pharr2016physically} and null-collision \cite{woodcock1965techniques}, also with its advanced variants \cite{kutz2017spectral}. Then based on the multiple importance sampling (MIS) \cite{veach1995optimally}, different extensions are also proposed for faster rendering speed \cite{nimier2022unbiased,lin2021fast}. Additionally, some adaptive methods are also showing an improvement in the performances \cite{ertl2007adaptive,wang2016parallel,hernell2009local}. These sampling strategies of classic explicit methods inspire us in the implicit neural field sampling, in spite of the high memory cost. Actually, this sampling scheme has a strong correlation to the neural radiance field's performances. The default way of Vanilla NeRF is to uniformly sample points between near and far planes. For forward-facing scenes like LLFF, NeRF proposes to transform rays from a truncated pyramid frustum into a cube, called normalized device coordinates (NDC), and sample uniformly in NDC space, which is equal to a linear sampling in disparity to infinity in the world space. This approach only fits planar light fields, meaning that all samples lie in front of the central camera frame. Many works borrowed the idea of NDC, and proposed a space-warping procedure to contract distant points into a bounded box. DoNeRF \cite{neff2021donerf} uses a logarithmic non-linearity to achieve non-linear sample placement. AdaNeRF \cite{kurz2022adanerf} tries to construct an adaptive sampling method to reduce the point sampling in one ray in a coarse-to-fine scheme. These methods often focus on the outside-looking-in scenes but are often not so efficient for large outdoor scenes.
Some methods like \cite{lin2021efficient,hu2022efficientnerf} take advantage of some prior knowledge to derive an efficient sampling but depth is often not accurate in the unbounded scenarios. 
Mip-NeRF-360 \cite{barron2022mip} introduces a contract function to map coordinates into a bounded ball. Some other similar reparameterization is also extended by \cite{isaac2022exact}. For the sake of saving computation cost, MeRF \cite{reiser2023merf} evolved the contraction to be piecewise-projective. But this contract is not often suitable for both foreground and background, making one part degrade dramatically. Our proposed method can fit the foreground and background separately to adaptively fit the different characteristics of the two regions.

\begin{figure*}[htbp]
\centering
\includegraphics[width=0.96\textwidth]{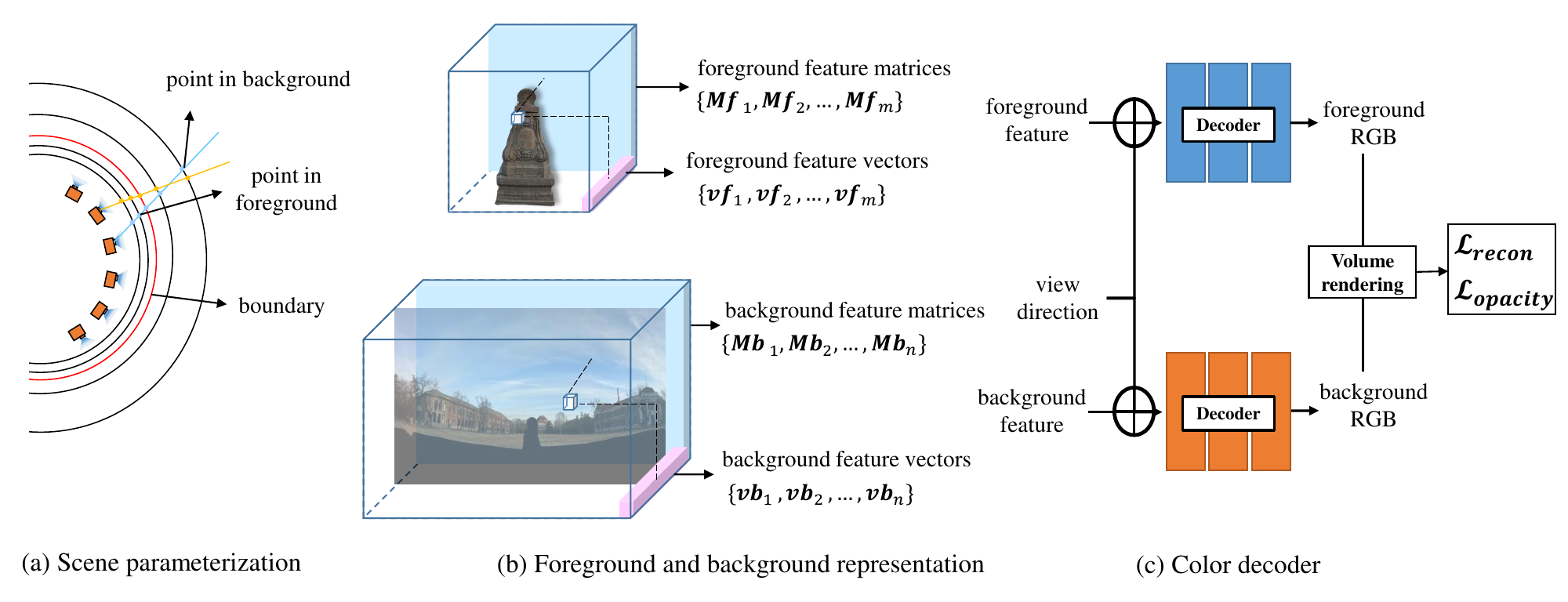}
\caption{\textbf{Overview of ImmersiveNeRF.} (a) Given a set of images captured by cameras placed on a sphere, we parameterize points in 3D space on multiple spherical shells; Meanwhile, the boundary of foreground and background shells is assigned to parameterize near and far points respectively.  (b) Given a sample point in our parameterized space, we keep a decomposed feature volume indicated as feature matrices perpendicular to the radius and feature vectors parallel to the radius to represent voxel features. (c) We decode the concatenation of the feature vector and view direction of a sampled point into its color, and perform volume rendering for multiple sampled points along a ray to get the pixel color.}
\label{figure:overview}
\end{figure*}

\section{Methods}
In this section, we introduce the process of ImmersiveNeRF in detail. 
We first propose the hybrid implicit representation for large space inside-looking-out immersive light field reconstruction, in which the foreground and the background are represented individually.
After that, we introduce a regularizer as an efficient constraint for the separation of foreground and background.  Finally, we present the depth adaption sampling strategy for the different parts to improve learning efficiency.
We show the paradigm of our method in Fig.~\ref{figure:overview}.
\subsection{Preliminaries}
As a highly efficient novel view synthesis method, NeRF constructs an implicit correspondence between the camera poses and the pixels' value. Specifically, the pixel, reflecting the process of light propagation, can be approximated as the integral of discrete sampling along the ray. For each point $\boldsymbol{p}$ in 3D space, given the observing direction $\boldsymbol{d}$, there is an unique volume density $\sigma(\boldsymbol{p})$ and color $c(\boldsymbol{p},\boldsymbol{d})$. NeRF estimates the density and color for each point with a simple MLP, which is defined as
\begin{equation}
\hat{\sigma}(\boldsymbol{p}),\hat{\boldsymbol{c}}(\boldsymbol{p},\boldsymbol{d}) = f(\boldsymbol{p},\boldsymbol{d})
\end{equation}
where $f(\cdot)$ denotes the MLP inference.
The pixel color is finally computed by accumulating a set of sample points on the ray via volume rendering. We denote the ray origin and direction as $\boldsymbol{o}$ and $\boldsymbol{d}$. Then the sampled points can be represented as $\boldsymbol{r}(t)=\boldsymbol{o}+t\boldsymbol{d}$, and the expected color $\hat{\boldsymbol{C}}(\boldsymbol{r})$ is computed by
\begin{equation}
    \hat{{C}}(\boldsymbol{r})=\int_{t_n}^{t_f} e^{-{\int_{t_n}^{t}\hat{\sigma}({\boldsymbol{r}}(s))}ds}
    \hat{\sigma}(\boldsymbol{r}(t))\hat{{c}}(\boldsymbol{r}(t))dt
\end{equation}
where $t_n,t_f$ are near and far bounds along a ray. 

We define all the rays formulating a ray set $R$. Given the ground truth value of pixels corresponding to the ray $C_{gt}(r)$, the loss of NeRF is determined by the MSE between $C_{gt}(r)$ and the estimation from Eq.(2), which is denoted as
\begin{equation}
L = \sum_{r\in R}\|\hat{\boldsymbol{C}}(\boldsymbol{r}) -C_{gt}(\boldsymbol{r})\|^2_2.
\end{equation}

The implicit representation in NeRF can be time-consuming for training and rendering.  To remedy this issue, hybrid representations are proposed by employing the advantages of both explicit and implicit representations. Any point in 3D space can be encoded by high-frequency features, which are stored explicitly in the form of a sparse voxel gird \cite{sun2022direct}, a hash table \cite{muller2022instant}, or a set of basis vectors or matrices with tri-plane decomposition \cite{chen2022tensorf}, and the features can be decoded via an implicit function, which can be an MLP \cite{tancik2020fourier}, or spherical harmonics \cite{fridovich2022plenoxels}. 
Different from the above research, in the next subsection, we propose a hybrid implicit representation based on the representation above to fit the large-scale inside-looking-out novel view synthesis.

\subsection{Hybrid scene representation}
The hybrid implicit representation is introduced first to cover the unbounded scene, which is referred to as NeRF++ \cite{zhang2020nerf++}.  We represent the whole large scene into two separate neural radiance fields, corresponding to the foreground radiance field and the background radiance field. For each part, we introduce a spherical transformation to achieve a compact representation conforming to the inside-looking-out task. 

\textbf{Scene Parameterization.}
Different from the typical scene in TensoRF \cite{chen2022tensorf}, the inside-looking-out task cannot be regarded as a cubic space to generate standard VM decomposition. Instead, we introduce a spherical-based decomposition to compactly express the inside-looking-out scene. We consider each observing standpoint as a hemisphere. For each point $\bold{p}=(x,y,z)$ in the scene, we find the nearest hemisphere of the  querying ray
and re-parameterize sample points in spherical space. All cameras are normalized accordingly to be concentric with the unit sphere. 
The foreground space and background space are mapped with different strategies. 

We first define the feature map as a function $\mathbf{\psi}$, for each point $\bold{p}$,  $\mathbf{\psi}(\bold{p})$ is generated by
\begin{equation}
    \mathbf{\psi}(\bold{p})=
    \left\{
            \begin{array}{lcccccl}
            (x, y, z),  & & \bold{p} \text{ in foreground}    &  \\
            (\theta, \phi, 1/r), & &   \bold{p}\text{ in background}   & 
            \end{array}
    \right.
\end{equation}
where $(\theta,\phi, r)$ is the spherical coordinates.

Then we construct $K$ factored matrices and vectors to represent density and appearance features for voxels in regular 3D space and use an MLP to decode the appearance features for the foreground region and the background region, respectively. The factored matrices and vectors of foreground are defined as $Mf_1(\cdot),..., Mf_K(\cdot)$ and $vf_1(\cdot),...,vf_K(\cdot)$. The factored matrices and vectors of background are defined as $Mb_1(\cdot),..., Mb_K(\cdot)$ and $vb_1(\cdot),...,vb_K(\cdot)$. Furthermore, the MLPs of foreground and background are determined as $f_f$ and $f_b$.
The estimated color and density for each point $\bold{p}$ at the foreground region are defined as 
\begin{equation}
\hat{\sigma}_f(\boldsymbol{p}),\hat{{c}}_f(\boldsymbol{p}) = f_f(\sum_{k=1}^K Mf_k(\mathbf{\psi}(\bold{p})) \cdot vf_k(\mathbf{\psi}(\bold{p})))
\end{equation}
and the estimated color and density for each point at the background  region are defined as 
\begin{equation}
\hat{\sigma}_b(\boldsymbol{p}),\hat{{c}}_b(\boldsymbol{p}) = f_b(\sum_{k=1}^K Mb_k(\mathbf{\psi}(\bold{p})) \cdot vb_k(\mathbf{\psi}(\bold{p})))
\end{equation}



In the next subsection, we will give the details of how to separate the background and foreground. We also present the estimation of the pixel value if the separation is determined. 



\textbf{Separation of foreground and background.}
The boundary depth is of great significance for foreground and background joint learning. 
In fact, the separation of the foreground and background depends on the disparity variance in different observing views. According to the principle depth estimation of stereo vision\cite{schoenberger2016mvs}, the closer the object is to the cameras, the greater the disparity changes. Meanwhile, the infinite distance does not change the disparity under different camera observations. We distinguish the separation depth of the foreground and background based on the disparity. We first define the baseline of the two furthest but overlapping cameras in the setting as $b$. The focal length of the cameras is denoted as $f$. From the two cameras, we can easily derive the disparity $d$ of a 3D keypoint with corresponding information from SfM \cite{schoenberger2016sfm}. If we define the 3D keypoint is lying on the separation boundary.
The separation bound $t_B$ can be calculated as 
\begin{equation}
t_B = \frac{f\times b}{d}.
\end{equation}

For each point sampled in one ray, once the separation bound $t_B$ is determined, based on Eq.(5-6), we have the estimated color as

\begin{equation}
\begin{split}
\hat{\boldsymbol{C}}(\boldsymbol{r})=&\int_{t_n}^{t_B} e^{-{\int_{t_n}^{t}\hat{\sigma}_f({\boldsymbol{r}}(s))}ds}
    \hat{\sigma}_f(\boldsymbol{r}(t))\hat{\boldsymbol{c}}_f(\boldsymbol{r}(t),\boldsymbol{d})dt \\
    +e^{-{\int_{t_n}^{t_B}\hat{\sigma}_f({\boldsymbol{r}}(s))}ds}&\int_{t_B}^{t_f} e^{-{\int_{t_B}^{t}\hat{\sigma}_b({\boldsymbol{r}}(s))}ds}
    \hat{\sigma}_b(\boldsymbol{r}(t))\hat{\boldsymbol{c}}_b(\boldsymbol{r}(t),\boldsymbol{d})dt.
    \end{split}
\end{equation}

The loss of color is defined the same as Eq. (3).



\subsection{Regularization for foreground and background separation}
Due to ill-posedness, NeRFs tend to overfit, which fit well in train views but collapse in test and novel views, exhibiting semi-transparent floating things in the air, leading to a severe failure for novel view rendering. This artifact called "floaters" in \cite{barron2022mip} is more noticeable when training views are sparse in the unbounded scenes. In our foreground and background separation setting, rays are cast for all pixels when training the foreground and background network. For little or no parallax background pixels, the foreground network may confuse with their depth, and thus generate floaters where there should be background pixels at the foreground field. In that case, points in both foreground and background space contribute to the final composite color in the background pixel. We will observe an incorrect semi-transparent object in the foreground when the rendering viewpoint slightly deviates from train views.
covering the corresponding object in the background.

We leverage the intuition that the foreground and the background pixels should be completely separate. In other words, the accumulated opacity along a ray in the foreground should either be one or zero, corresponding to the pixel of either background content or foreground content. More specifically, we estimate the transmittance $T(\boldsymbol{r})$ for the foreground:
\begin{equation}
    T(\boldsymbol{r}) = e^{-\int_{t_n}^{t_B}\sigma_f({\boldsymbol{p}(s))}ds}.
\end{equation}
Then we can construct a BCE loss for this $T(\boldsymbol{r})$ to constraint it to be either one or zero,
\begin{equation}
    \boldsymbol{L_{\textit{reg}}}(\boldsymbol{r})=-T(\boldsymbol{r})\log{T(\boldsymbol{r})} - (1-T(\boldsymbol{r}))\log(1-T(\boldsymbol{r})).
\end{equation}
With this opacity loss, we can separate the density in the foreground and background more clearly and the floaters in the foreground covering the background will be constrained to zero transmittance with low density. We demonstrate the effectiveness of our regularizer in the ablation study for reducing the floaters in novel views.



\subsection{Sampling strategy}

Sampling strategy is essential for the foreground and background joint training. The hybrid representation functions well when the two networks iterate at a balanced rate. Hence, we also design a corresponding sampling strategy for our hybrid scene representation. 

\begin{figure}[htbp]
\centering
\includegraphics[width=0.25\textwidth]{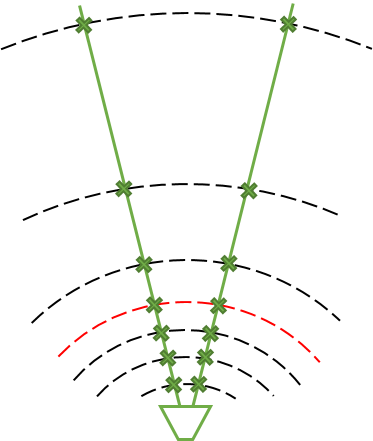}
\captionsetup{justification=centering}
\caption{Sampling Strategy. Uniform samples in equal steps of depth in the foreground (below the red dotted line). Uniform samples in equal steps of disparity in the background (above the red dotted line).}
\label{figure:sample}
\end{figure}

In our reparameterized space, Fig. \ref{figure:sample} illustrates that points are sampled along the radius of multiple concentric spherical shells. For the foreground scene, we take uniform samples for $t$. For the background scene, we also take uniform samples but in inverse radius $1/r$. Then we compute the coordinate of the intersection point between a ray and a spherical shell with the radius $r$ for the background. 
We also take the coarse-to-fine sampling strategy. At the beginning of the training, we sample uniform sparse points, and the appearance feature is also in low resolution correspondingly. Then the appearance feature vectors and matrices are upsampled linearly and bilinearly, and so is the number of sample points. The resolutions are different for the foreground and background to adapt to the different characteristics. Some perturbs for the sample points are also conducted to avoid overfitting. Experiments show that our proposed strategy can effectively alleviate the floater problem. 
\begin{figure*}[htbp]
\centering
\begin{tabular}{ccc}
\begin{tabular}{c}
    \includegraphics[height=0.245\linewidth]{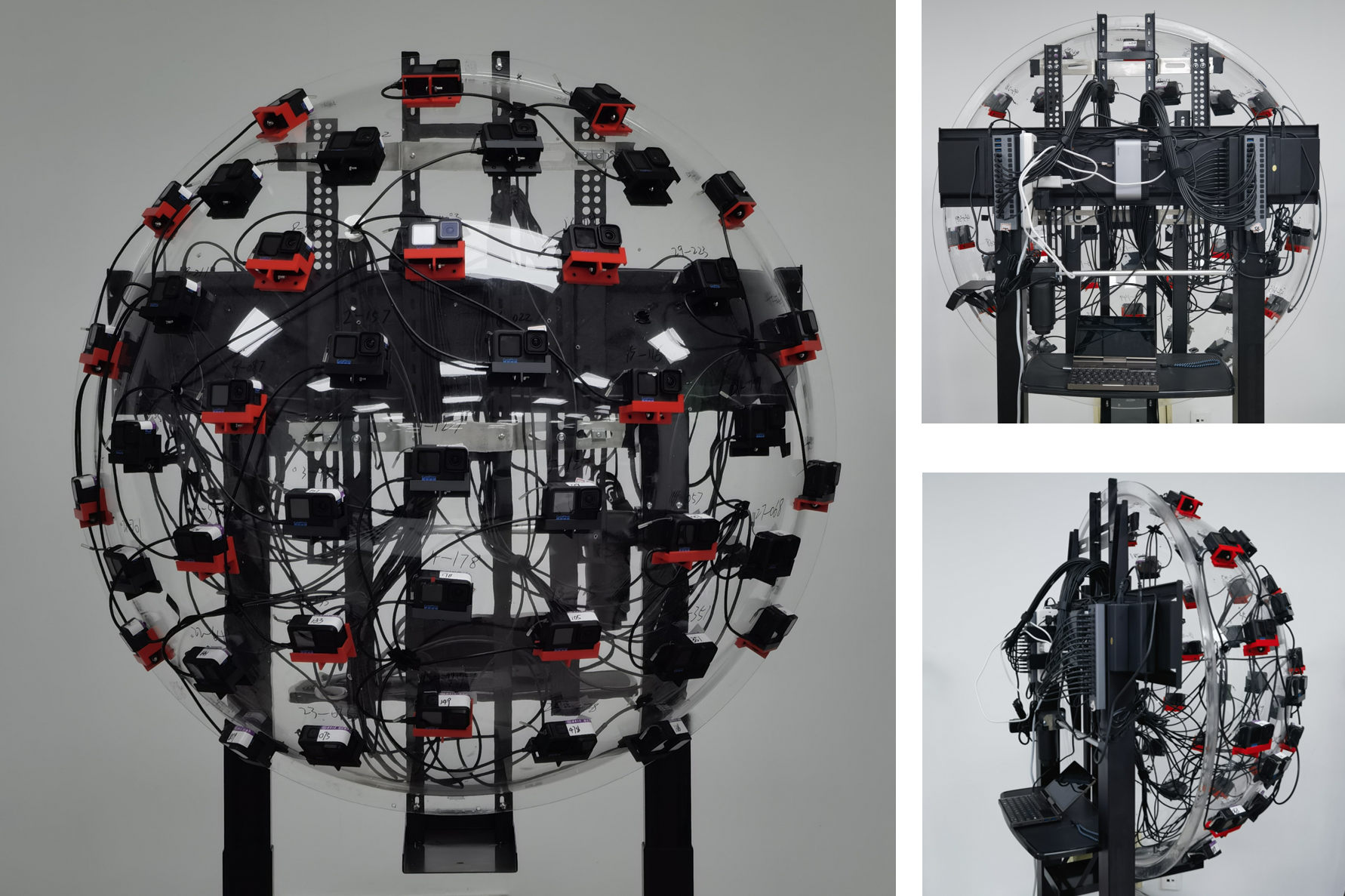} 
    \\(a) Capture system
\end{tabular}&
\begin{tabular}{c}
    \includegraphics[height=0.245\linewidth]{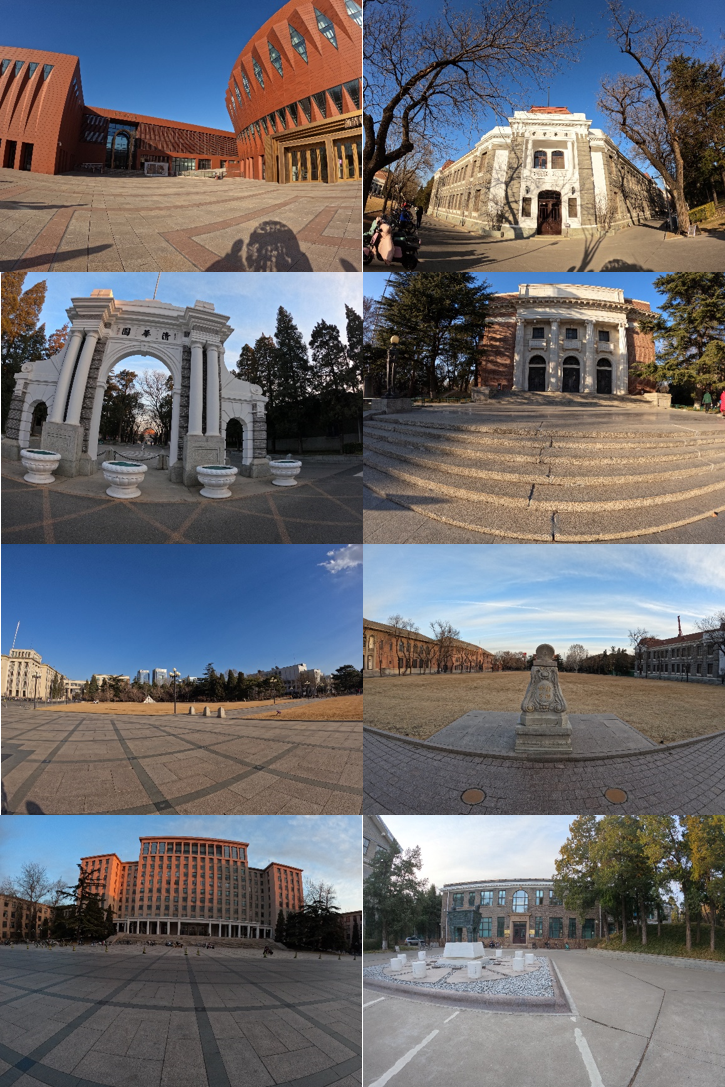} 
    \\(b) Gallery
\end{tabular}&
\begin{tabular}{c}
    \includegraphics[height=0.245\linewidth]{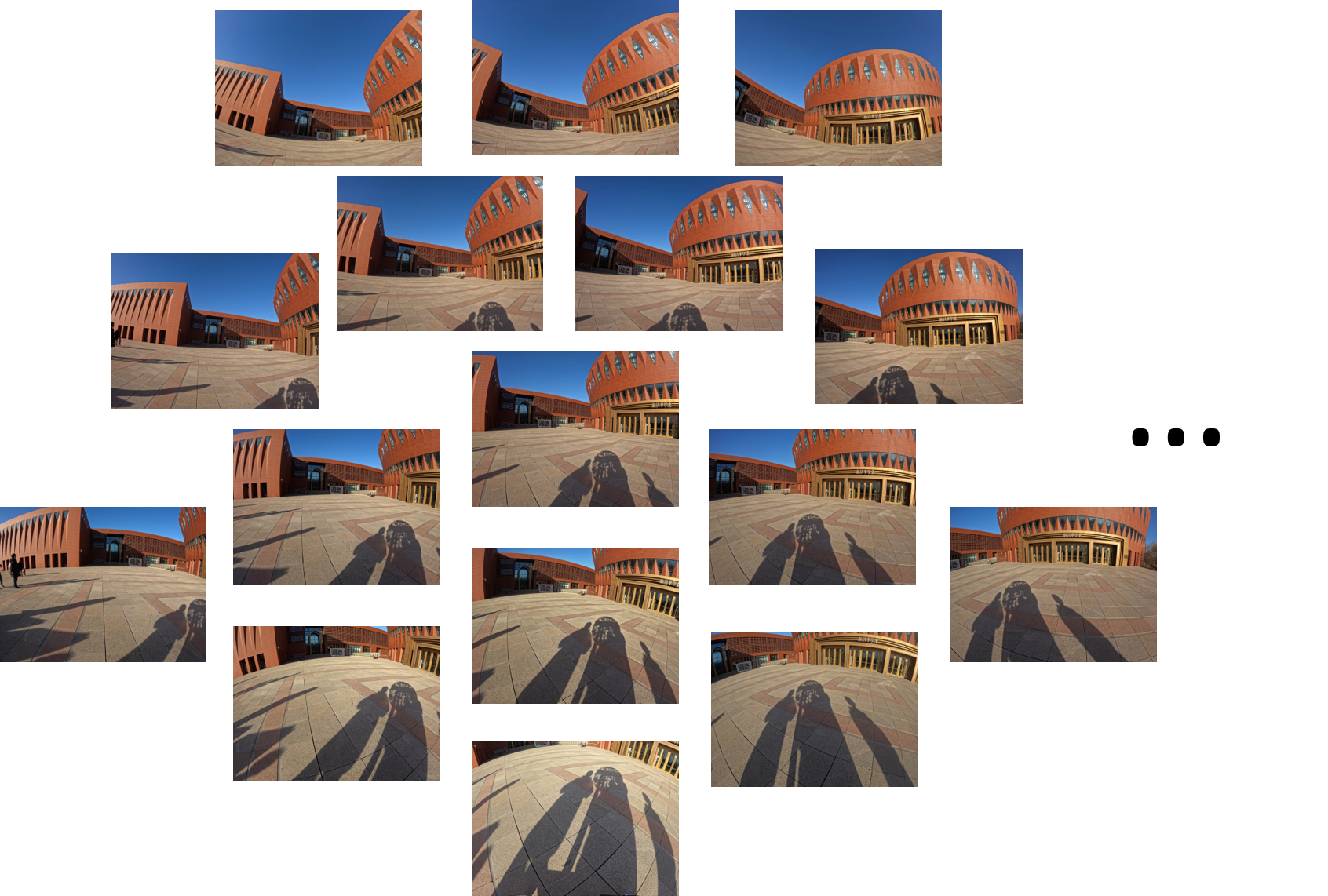} 
    \\(c) Real captured data
\end{tabular} 
\end{tabular}
\caption{Capture Settings.}
\label{figure:system}
\end{figure*}

\section{Experiments}
\subsection{Evaluation settings}
\subsubsection{Dataset}
\textbf{THUImmersive Dataset.}
Traditional multi-view images datasets are mainly captured in an outside-looking-in manner, such as LLFF \cite{mildenhall2019local}, NeRF blender datasets \cite{mildenhall2021nerf}, and Mip-NeRF-360 \cite{barron2022mip} dataset, aiming at the reconstruction and rendering of a single object or a planar light field. In order to fulfill immersive novel view rendering in a large-scale scene, we captured the outdoor scene in an inside-looking-out manner. 
To capture large outdoor scenes in an efficient way, we build our mobile multi-view capture system following the camera layout in \cite{broxton2020immersive}. Our capture system is composed of 42 GoPro Black Hero 10 cameras placed at the vertices of a v3 icosahedral tiling of a 1-meter diameter hemisphere. Fig. \ref{figure:system} shows our multi-view capture setup in reality. We capture images synchronously at a resolution of $5568\times4176$ in a fisheye projection manner. We obtain the initial camera intrinsic from each camera's EXIF file and calibrate camera intrinsic and extrinsic parameters using Metashape. We treat the central view as a test view for qualitative and quantitative evaluation, and the others are used for training.

As shown in Fig. \ref{figure:system} (b), we collected 8 outdoor scenes on Tsinghua campus. Most scenes contain landmarks, such as the Great Hall, the THUGate, and the sundial. For most scenes in our dataset, the depth range is large, including nearby objects, far-away buildings, and infinite backgrounds. Due to connection and synchronization failures, each scene contains about 40 views evenly distributed on the sphere. Furthermore, to enlarge the movable range of viewpoints, we also include multiple neighboring viewpoints for the same scene. We translate the 
camera rig laterally with the same orientation, broadening the baseline of overall cameras to more than 5 meters. In that case, each scene contains more than 200 views.

\textbf{Immersive Light Field Dataset.} 
We also demonstrate the generality of our method using synchronized multi-view images from some large-scale scenes, such as Goats, Dog, and Car in the immersive light field dataset. 

\begin{figure*}[htb]
\centering
\includegraphics[width=0.85\textwidth]{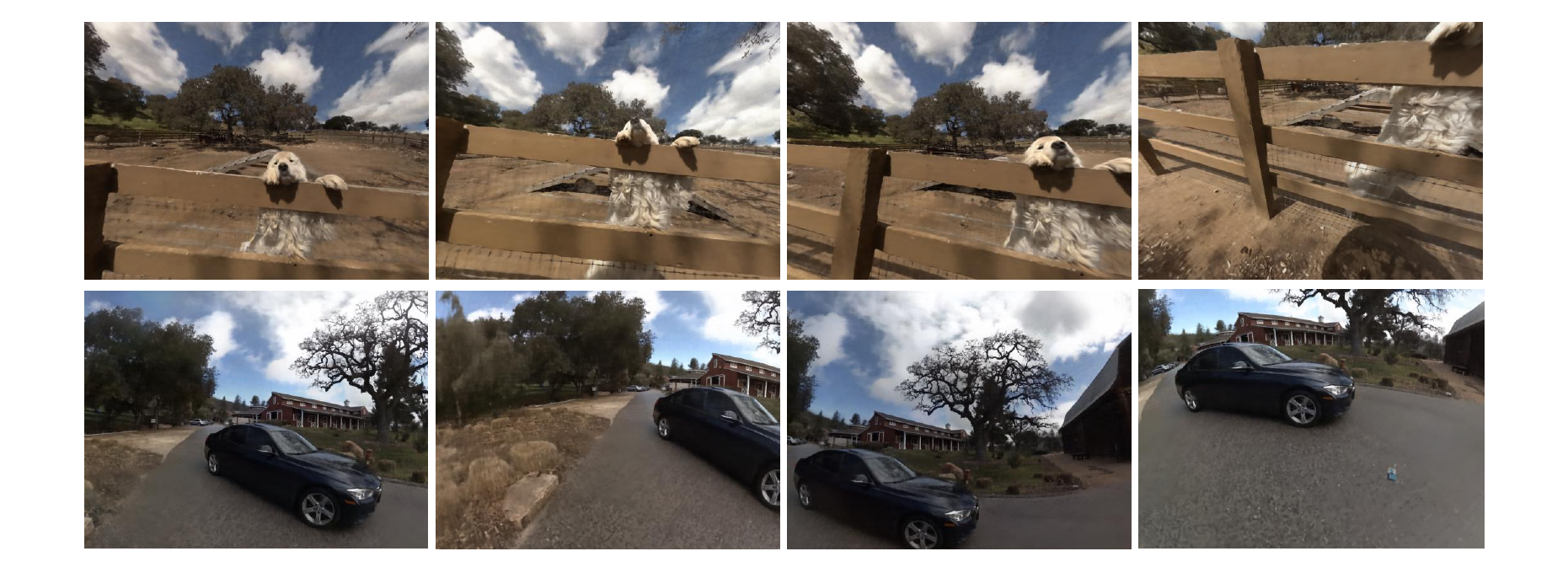}
\caption{Snapshots of rendered views on Immersive Light
Field dataset.}
\label{figure:immersivedataset}
\end{figure*}

\begin{figure*}[htb]
\centering
\includegraphics[width=0.85\textwidth]{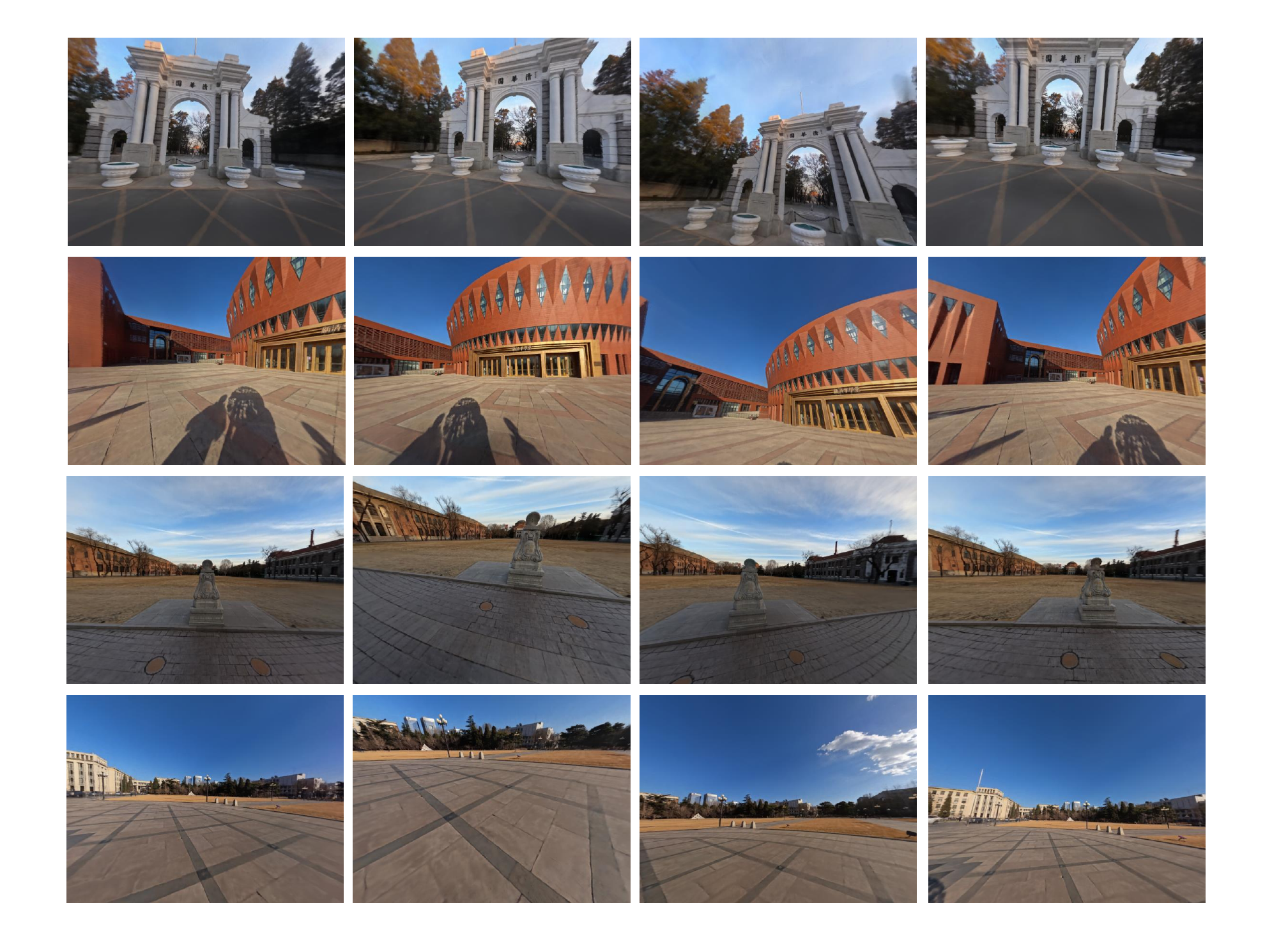}
\caption{Snapshots of rendered novel views on THUImmersive dataset.}
\label{figure:ourdataset}
\end{figure*}

\subsubsection{Baselines}
We compare our methods to the following baselines:
\begin{itemize}
    \item \textbf{Mip-NeRF-360 \cite{barron2022mip}:} A neural rendering framework focusing on the unbounded scenes with well-designed ray parameterization and sampling. 
    \item \textbf{Nerf++ \cite{zhang2020nerf++}:} A framework based on the NeRF with simple foreground and background decomposition modeling.
    \item \textbf{IBRNet \cite{wang2021ibrnet}:} A learnable rendering scheme based on the interpolation of the neighbor viewpoints.
\end{itemize}   

\begin{figure*}[htbp]
\centering
\begin{tabular}{cccccc}
\begin{tabular}{c}
\hspace{-3em}
    \includegraphics[width=0.2\linewidth]{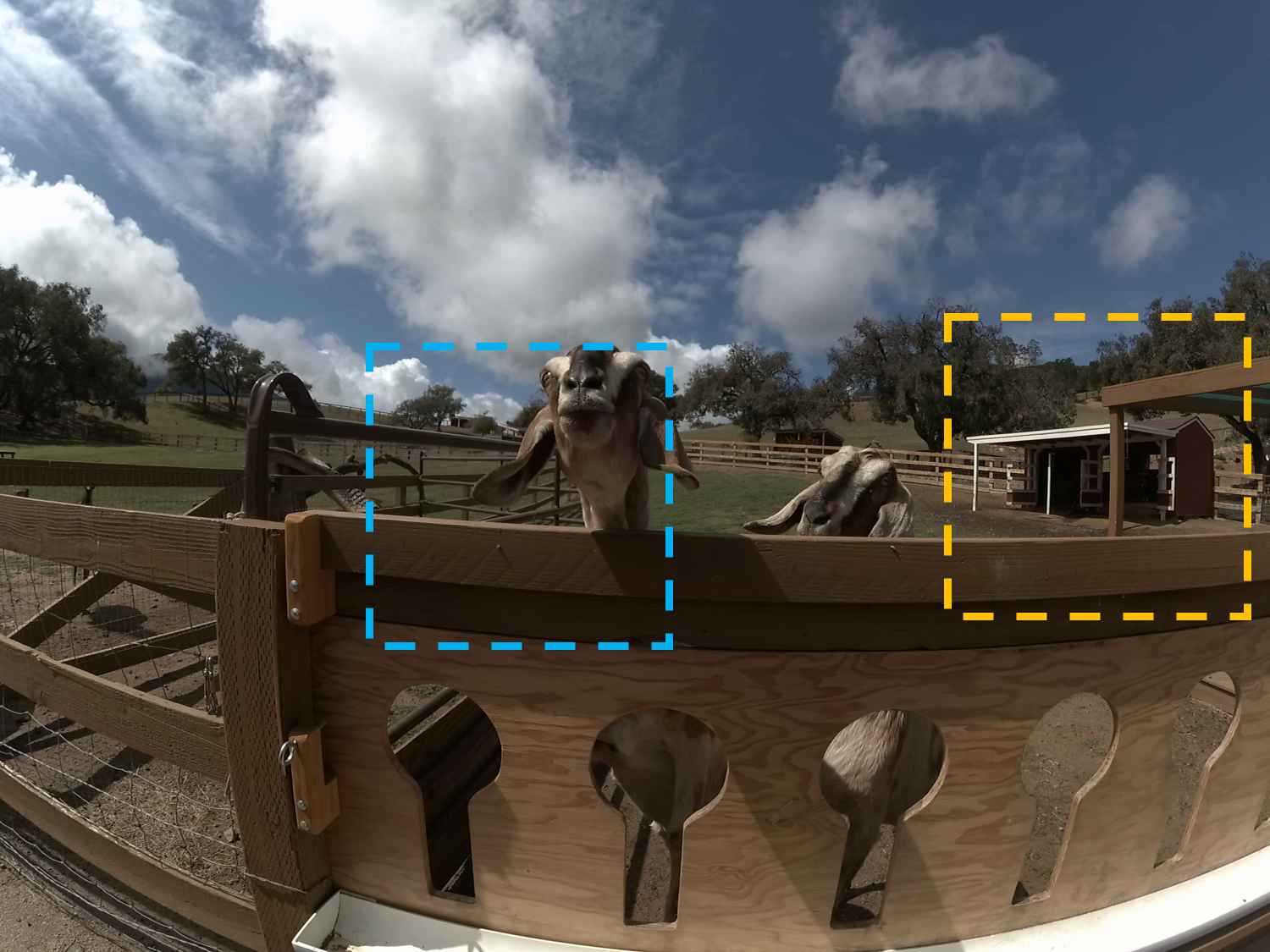} 
    \includegraphics[width=0.15\linewidth]{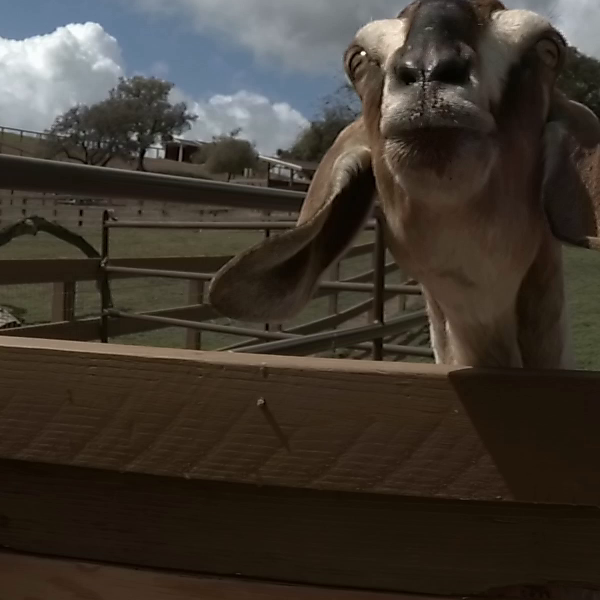} 
    \includegraphics[width=0.15\linewidth]{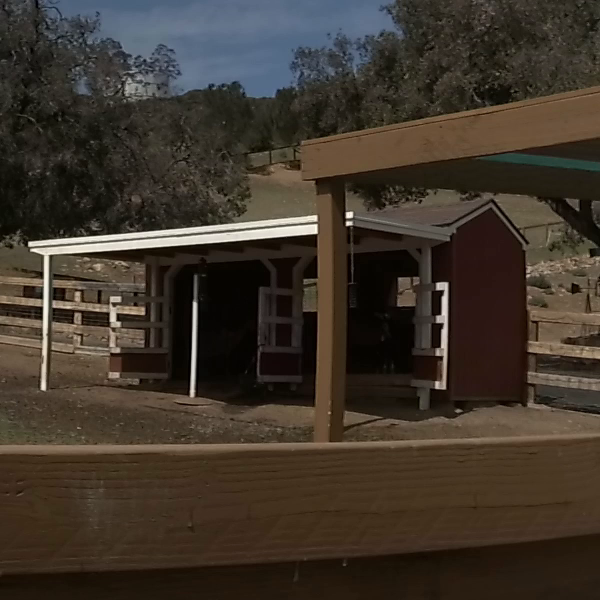}
    \\(a) Ground Truth
\end{tabular}&
\begin{tabular}{c}
\hspace{-2em}
    \includegraphics[width=0.2\linewidth]{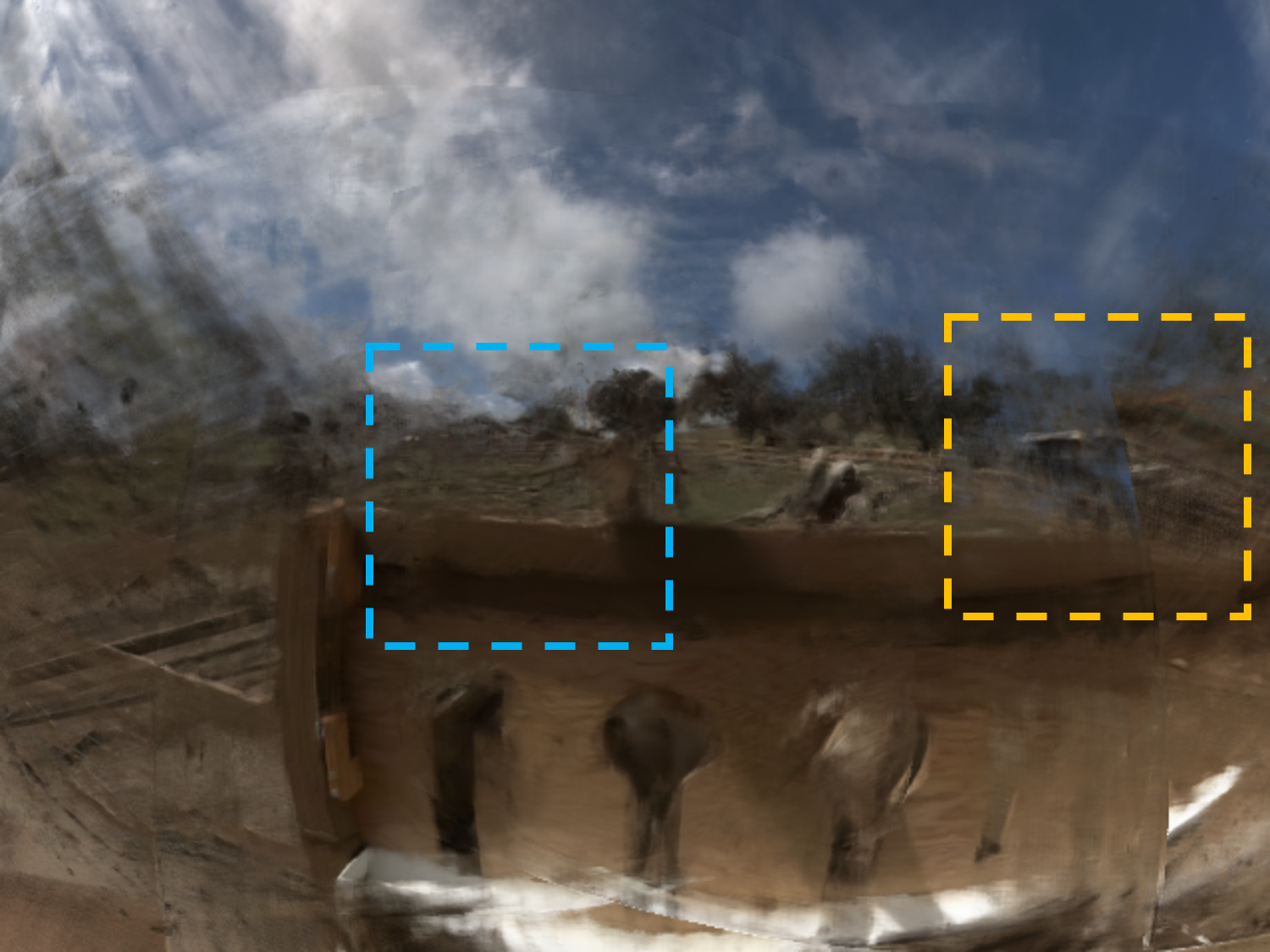} 
    \includegraphics[width=0.15\linewidth]{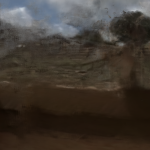}
    \includegraphics[width=0.15\linewidth]{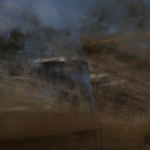}\\(b) IBRNet without fine-tune
\end{tabular}&
\\
\begin{tabular}{c}
\hspace{-3em}
    \includegraphics[width=0.2\linewidth]{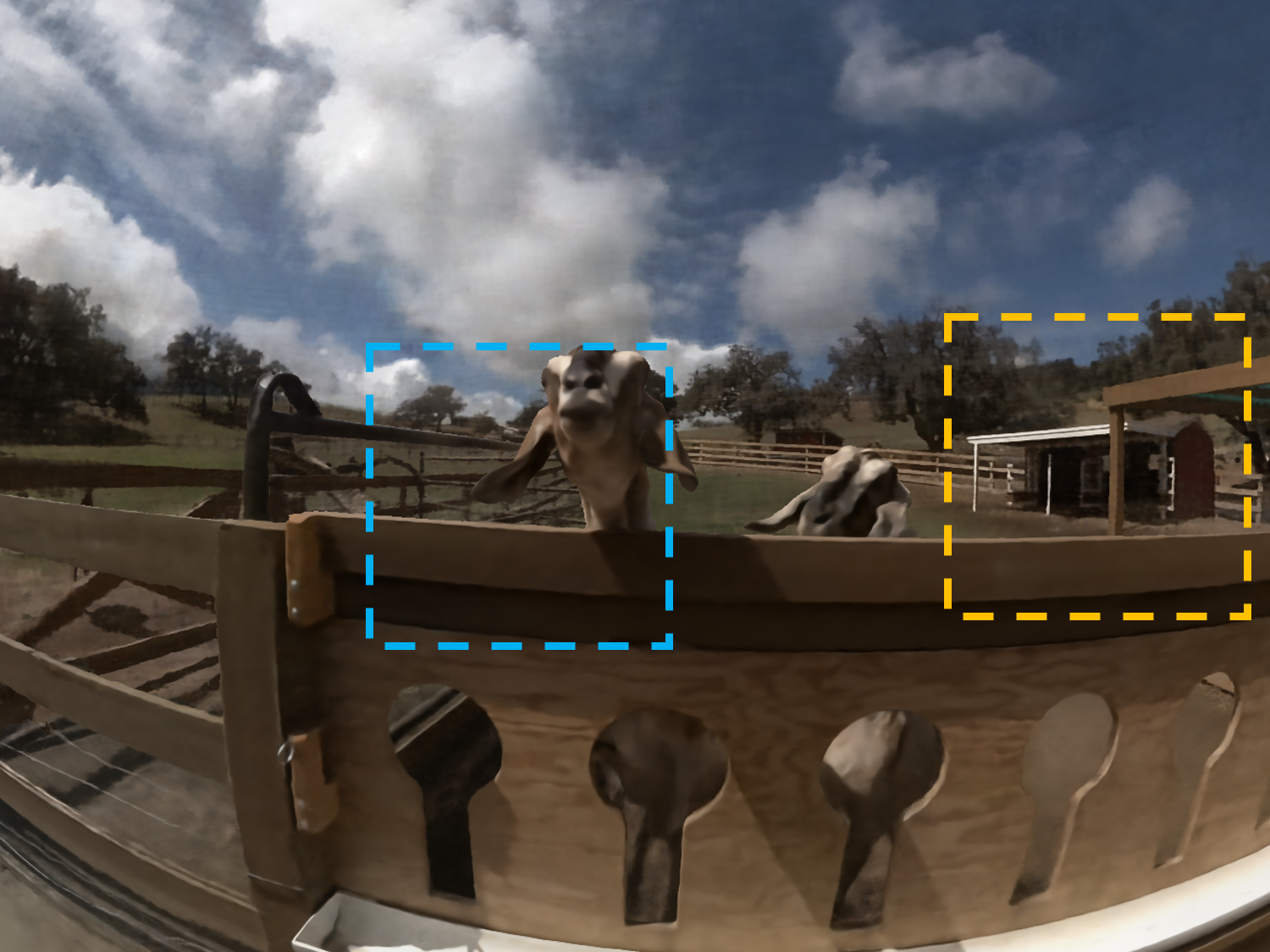} 
    \includegraphics[width=0.15\linewidth]{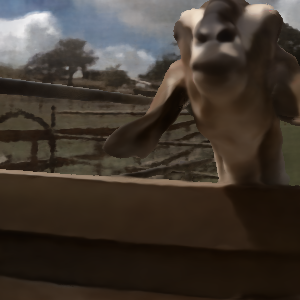} 
    \includegraphics[width=0.15\linewidth]{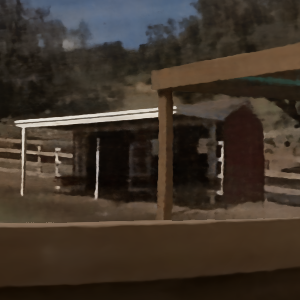}
    \\(c) Ours
\end{tabular}&
\begin{tabular}{c}
\hspace{-2em}
    \includegraphics[width=0.2\linewidth]{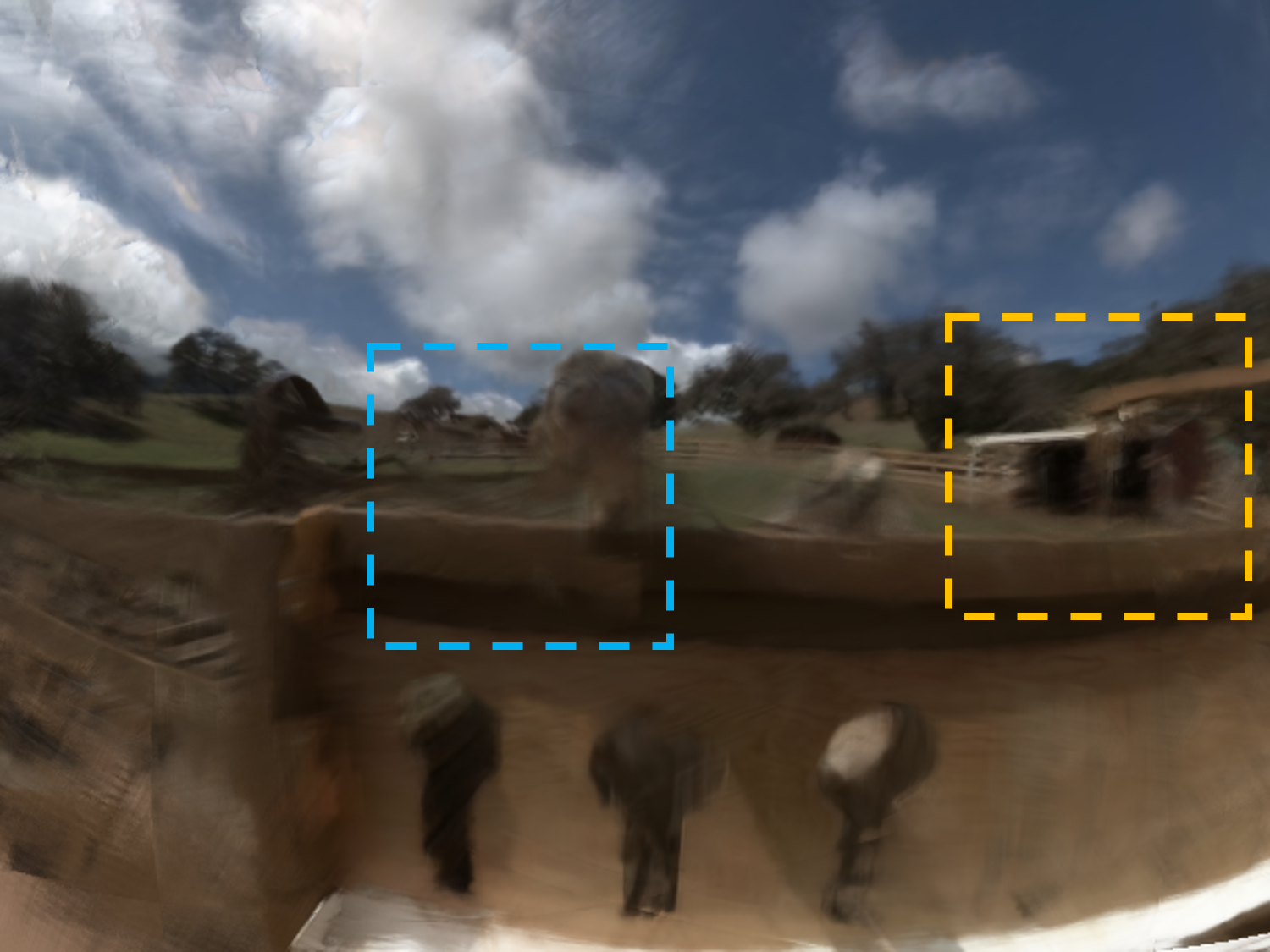} 
    \includegraphics[width=0.15\linewidth]{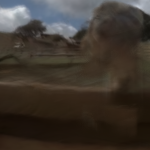}
    \includegraphics[width=0.15\linewidth]{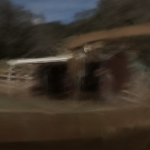}
    \\(d) IBRNet with fine-tune
\end{tabular}&
\\
\begin{tabular}{c}
\hspace{-3em}
    \includegraphics[width=0.2\linewidth]{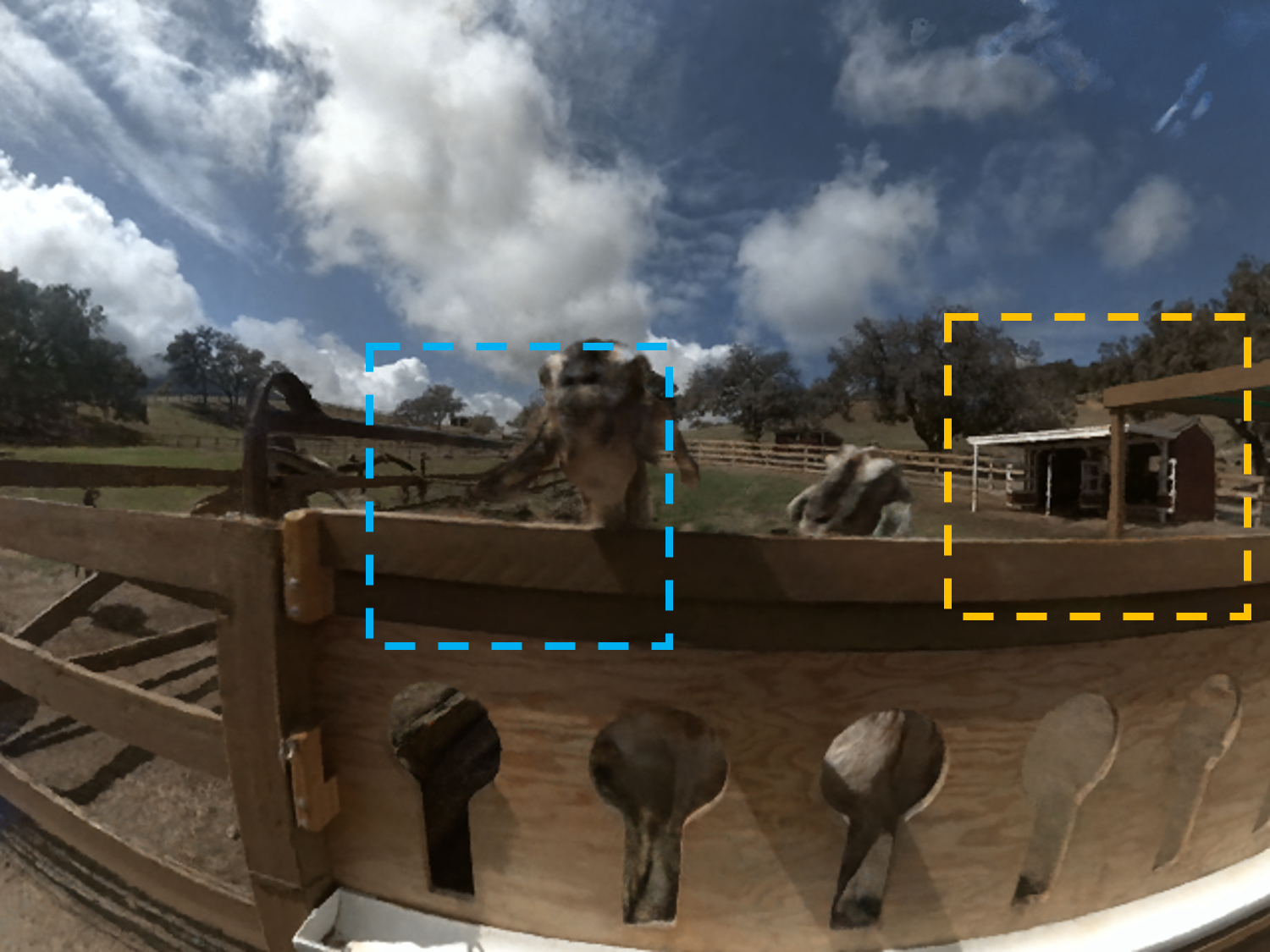} 
    \includegraphics[width=0.15\linewidth]{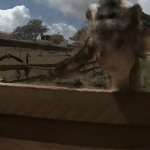}
    \includegraphics[width=0.15\linewidth]{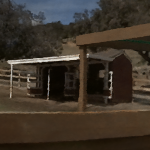}
    \\(e) Mip-NeRF-360
\end{tabular}&
\begin{tabular}{c}
\hspace{-2em}
    \includegraphics[width=0.2\linewidth]{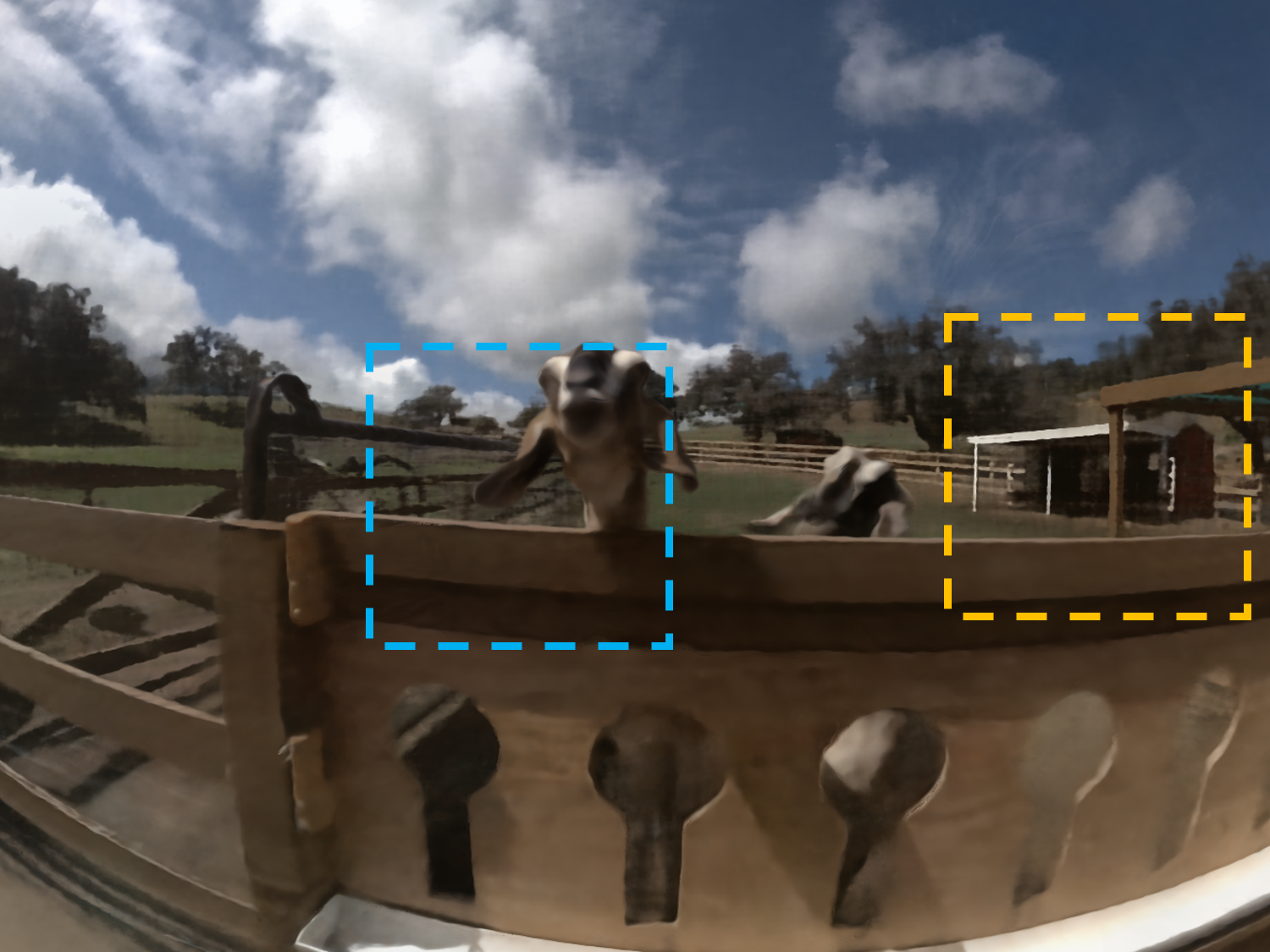} 
    \includegraphics[width=0.15\linewidth]{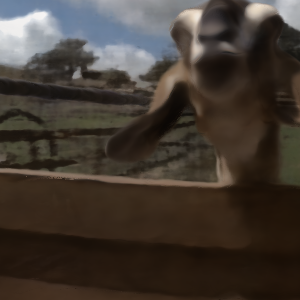}
    \includegraphics[width=0.15\linewidth]{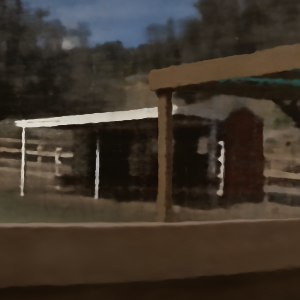}
    \\(f) NeRF++
\end{tabular}
\end{tabular}
\caption{Comparison of our model to existing methods on test view and zoom-in details.}
\label{figure:goasttest}
\end{figure*}

\textbf{Training details.} For consistent comparison, we re-implement fisheye projection for all methods. We treat the central fisheye view as the test view for evaluation and use the others for training, as shown in Fig. \ref{figure:goasttest} and Tab. \ref{table:quantitative}. We fine-tuned IBRNet for each scene for another 20,000 steps.
We optimize our model with a batch size of 4096 rays on a single NVIDIA GeForce 3090 GPU (24GB). The average optimization time is 48.7 minutes and the average rendering time is 42.8 seconds for 2k resolution.

\textbf{Metrics.} We evaluate the rendering quality on the central test view and the following quantitative metrics: (1) Peak signal-to-noise ratio (PSNR); (2) Structural dissimilarity index measure (SSIM); (3) Perceptual quality measure LPIPS.

\subsection{Results}

\begin{table*}[htbp]
\setlength{\tabcolsep}{0.95mm}  
\caption{Quantitative comparison of our proposed method to baselines of existing methods}
\begin{tabular}{c|ccccc|ccccc|ccccc}
\hline
\multirow{2}{*}{Methods} &
  \multicolumn{5}{c|}{PSNR$\uparrow$} &
  \multicolumn{5}{c|}{SSIM$\uparrow$} &
  \multicolumn{5}{c}{LPIPS$\downarrow$} \\ \cline{2-16} 
 &
  Goats &
  Dog &
  Car &
  Thugate &
  Sundial &
  Goats &
  Dog &
  Car &
  Thugate &
  Sundial &
  Goats &
  Dog &
  Car &
  Thugate &
  Sundial \\ \hline
Mip-NeRF-360 &
  25.23 &
  22.88 &
  23.00 &
  22.35 &
  23.76 &
  0.8054 &
  0.6813 &
  0.7558 &
  0.7126 &
  0.7255 &
  \textbf{0.1417} &
  0.3579 &
  0.2777 &
  \textbf{0.1886} &
  0.2355 \\
NeRF++ &
  26.60 &
  24.21 &
  20.01 &
  20.52 &
  22.51 &
  0.8029 &
  0.6600 &
  0.5877 &
  0.5675 &
  0.6360 &
  0.3160 &
  0.4281 &
  0.5208 &
  0.4625 &
  0.3524 \\
IBRNet(w/o finetune) &
  16.33 &
  15.22 &
  15.62 &
  15.48 &
  16.08 &
  0.5219 &
  0.3858 &
  0.4012 &
  0.4749 &
  0.5221 &
  0.5489 &
  0.6106 &
  0.5904 &
  0.5988 &
  0.6867 \\
IBRNet(w finetune) &
  22.58 &
  20.20 &
  19.90 &
  19.75 &
  20.95 &
  0.6679 &
  0.5136 &
  0.5239 &
  0.5713 &
  0.6037 &
  0.4515 &
  0.5674 &
  0.5572 &
  0.5460 &
  0.5911 \\
Ours &
  \textbf{28.16} &
  \textbf{25.82} &
  \textbf{23.56} &
  \textbf{23.93} &
  \textbf{24.22} &
  \textbf{0.8343} &
  \textbf{0.7266} &
  \textbf{0.7900} &
  \textbf{0.7589} &
  \textbf{0.7621} &
  0.2091 &
  \textbf{0.2389} &
  \textbf{0.2384} &
  0.2443 &
  \textbf{0.2021} \\ \hline
\end{tabular}

\label{table:quantitative}
\end{table*}

We demonstrate our novel view rendering results on different scenes in Fig. \ref{figure:teaser}. Our reconstructed model can enable near photorealistic rendering at 2K resolution. In the Supp. Video, we render the scene in a large baseline trajectory showing view-pleasant interpolation and extrapolation effects. 

Fig. \ref{figure:immersivedataset} and Fig. \ref{figure:ourdataset} show a few rendered novel views in the Immersive Light Field dataset and THUImmersive, respectively. Thanks to the hybrid representation and the opacity loss, our method behaves well on both view interpolation and view extrapolation with few artifacts. 

\subsubsection{Qualitative comparison}
We compare interpolated and extrapolated renderings with the baselines. We highlight visual comparisons on the test view of our methods to the baselines in Fig. \ref{figure:goasttest}. NeRF++ tends to render blurred images, especially in foreground regions. We take IBRNet pre-trained model to infer a novel view, it shows that IBRNet can fit training images perfectly, but collapses on the test. When fine-tuning IBRNet for each scene, it can also hardly fit well. 
Although Mip-NeRF-360 can render sharp visual effects in high-frequency textures, its floater artifacts rendered on the background are also noticeable, e.g., the colorful floating things before clouds and trees in the background. We also compare zoom-in details rendered on novel views. Zoom-in regions in the blue frame show that on some occluded thin structures, like the forward-extending fence, our method can learn its texture and depth most accurately, with the least ghosting things. And our method also exhibits the best visual effect on the goat's face. 
Zoom-in regions in the yellow frame contain both the far-away house and shed and the infinite sky. Mip-NeRF 360 exhibits sharp results on trees and sky at the cost of the blurry house and shed. And NeRF++ is on the contrary. It can be seen that our method generates finer details for all nearby goats, far-away house, and infinite trees and sky. This is because we utilize the spherical feature field with varied sampling strategies to better fit the high-frequency details.




\subsubsection{Quantitative comparison}
Tab. \ref{table:quantitative} shows the quantitative comparison of our methods to the baselines using PSNR, SSIM, and LPIPS of the test image. We demonstrate the effectiveness of our method on three unbounded scenes, Goats, Dog, and Car from the immersive light field dataset and two scenes, Thugate, and Sundial in our captured THUImmersive. Higher PSNR and SSIM indicate a better rendering quality, and for LPIPS, lower numbers indicate better quality. As Tab. \ref{table:quantitative} shows our method exhibits the highest PSNR and SSIM in all five scenes. The quantitative results illustrate the effectiveness of our methods compared with other baseline methods quantitatively.

\subsection{Ablation study}
\subsubsection{Effect of opacity loss}
\begin{figure*}[htbp]
\centering
\begin{tabular}{cc}
\begin{tabular}{c}
    \includegraphics[width=0.27\linewidth]{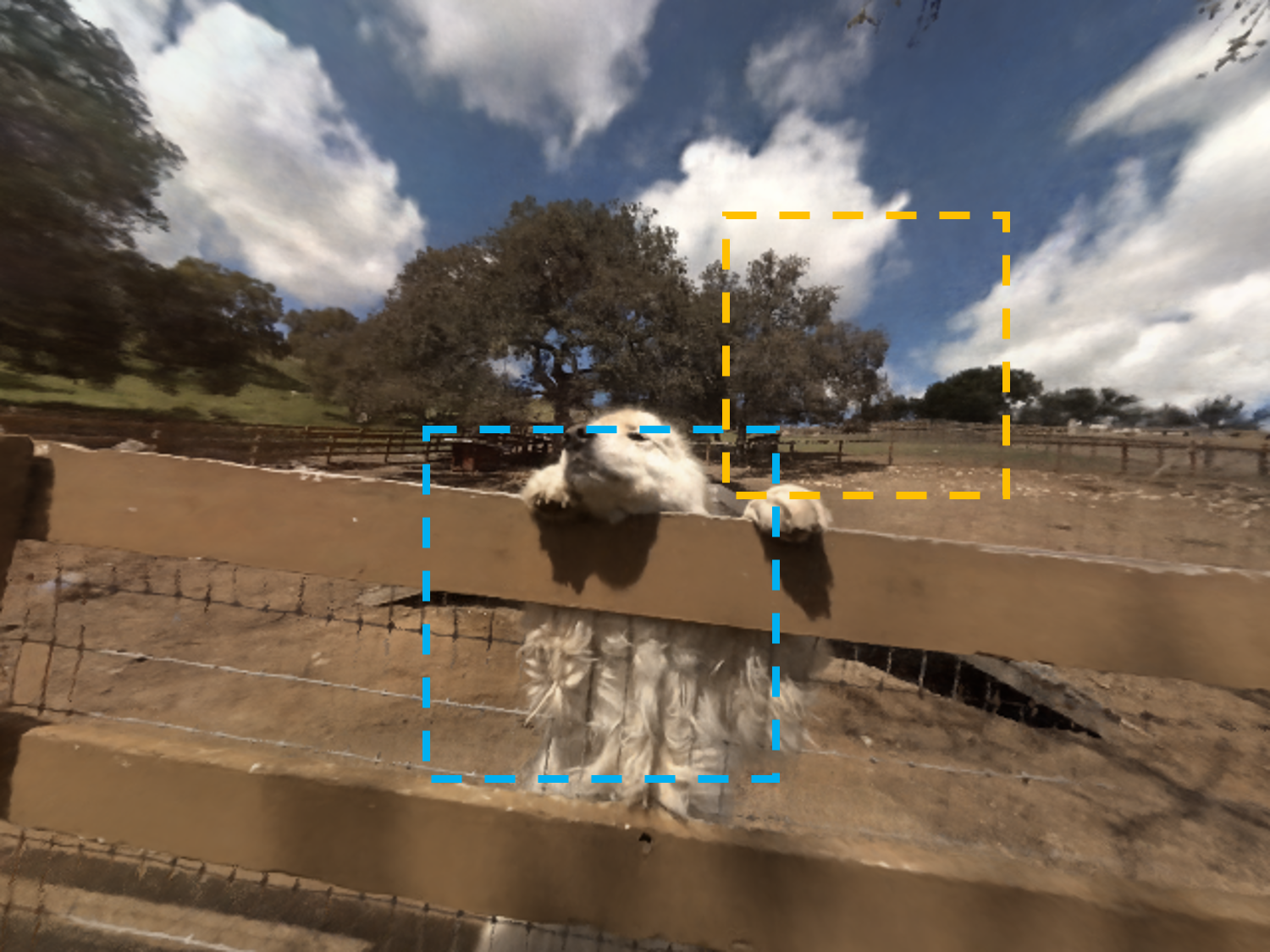}
    \includegraphics[width=0.2\linewidth]{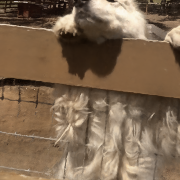}
    \includegraphics[width=0.2\linewidth]{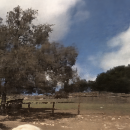}
    \includegraphics[width=0.27\linewidth]{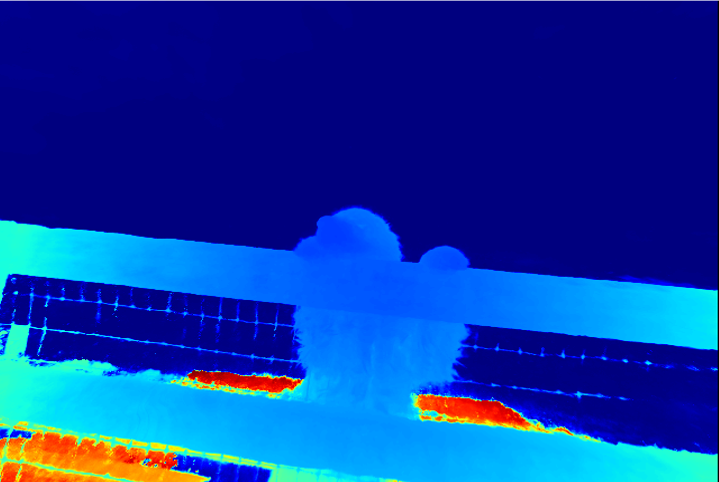} 
    \\(a) Novel view and depth map with opacity loss
\end{tabular}&
\\
\begin{tabular}{c}
    \includegraphics[width=0.27\linewidth]{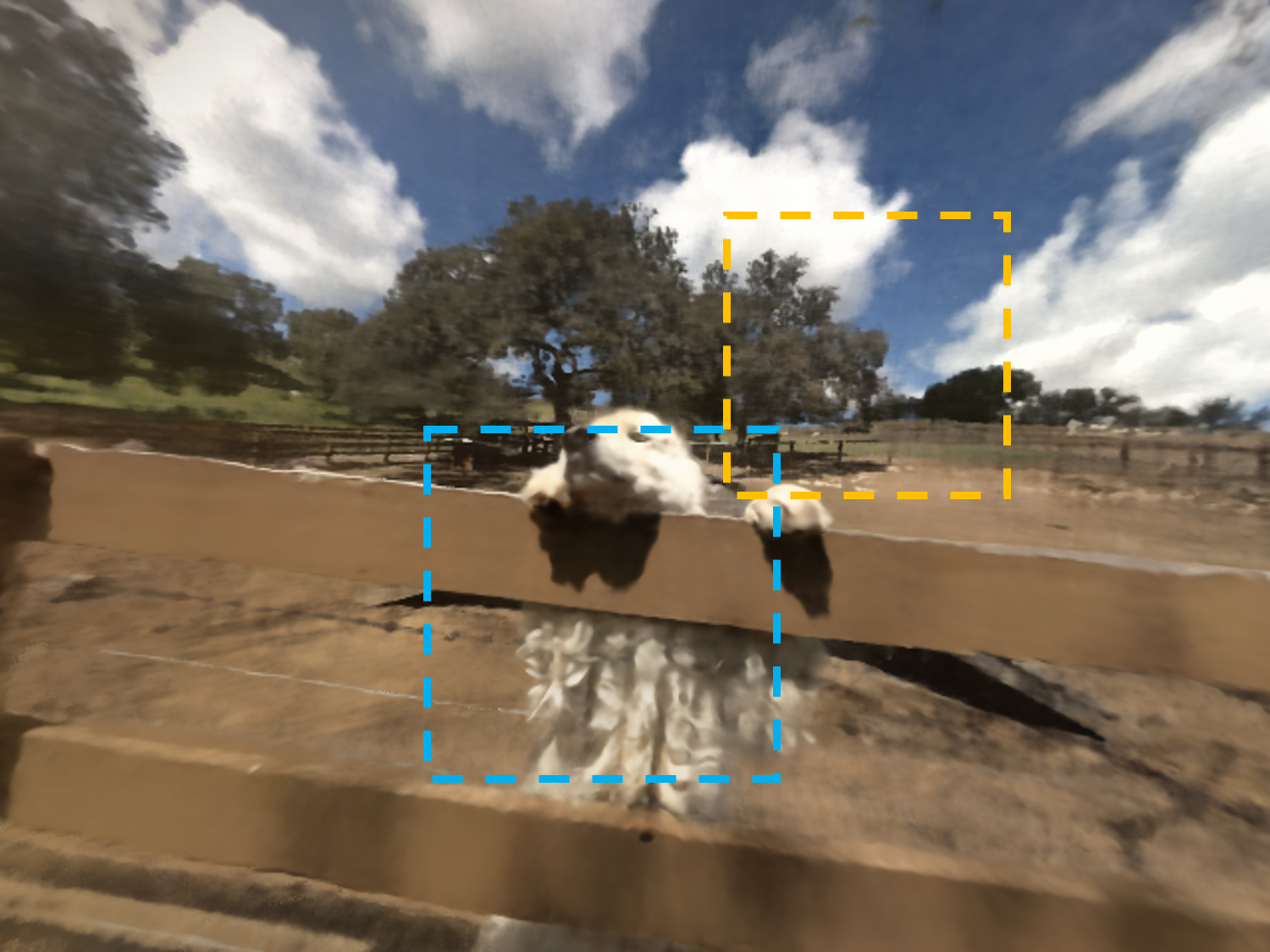}
    \includegraphics[width=0.2\linewidth]{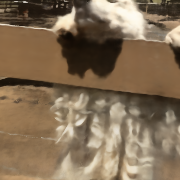}
    \includegraphics[width=0.2\linewidth]{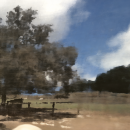}
    \includegraphics[width=0.27\linewidth]{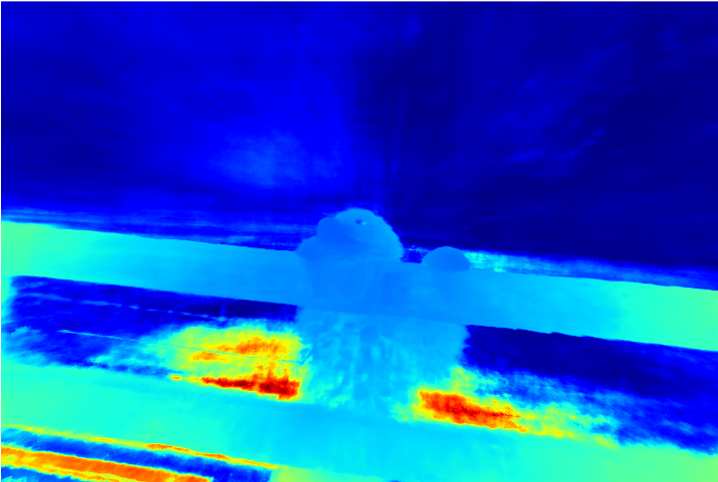}
    \\(b) Novel view and depth map without opacity loss
\end{tabular} 
\end{tabular}
\caption{Comparison on novel view and its foreground depth map.}
\label{figure:depth}
\end{figure*}

As mentioned in 3.3, we introduce opacity loss to alleviate the floaters artifacts. We ablate our method on the scene Dog with and without opacity loss and render a novel view and its foreground depth map, respectively, in Fig. \ref{figure:depth}. Fig. \ref{figure:depth} shows that the rendered depth map with opacity loss is more accurate and compact than that without opacity loss. In rendered foreground depth, the dark blue indicates that the disparity is so small that out of range of the assigned foreground region. The depth map in (a) exhibits a clear separation of the foreground and background. But for depth in (b), there are still some weights in the regions (like sky and tree pixels) where supposed to be rendered as background. And it is these weights that result in floater artifacts. Also, in (a) the depth of some thin structures, like the iron wire between handrails and the dog hairs, is more noticeable than that in (b). This can also be seen in the rendered view and zoom-in details. Table \ref{table: ablation} shows that opacity loss contributes to the overall rendering effect. 


\begin{table}[htb]
\centering
\caption{Opacity loss ablation.}
\begin{tabular}{ccccc}
\toprule
Scene                & Opacity loss & PSNR $\uparrow$ & SSIM $\uparrow$ & LPIPS $\downarrow$ \\
\midrule            
Dog               & Without      &  24.92    &   0.7072   &  0.2672    \\
                       & With         &  \textbf{25.82}    &   \textbf{0.7266}   &  \textbf{0.2389}    \\
\bottomrule
\end{tabular}
\label{table: ablation}
\end{table}


\begin{figure}[htb]
\centering
\includegraphics[width=0.48\textwidth]{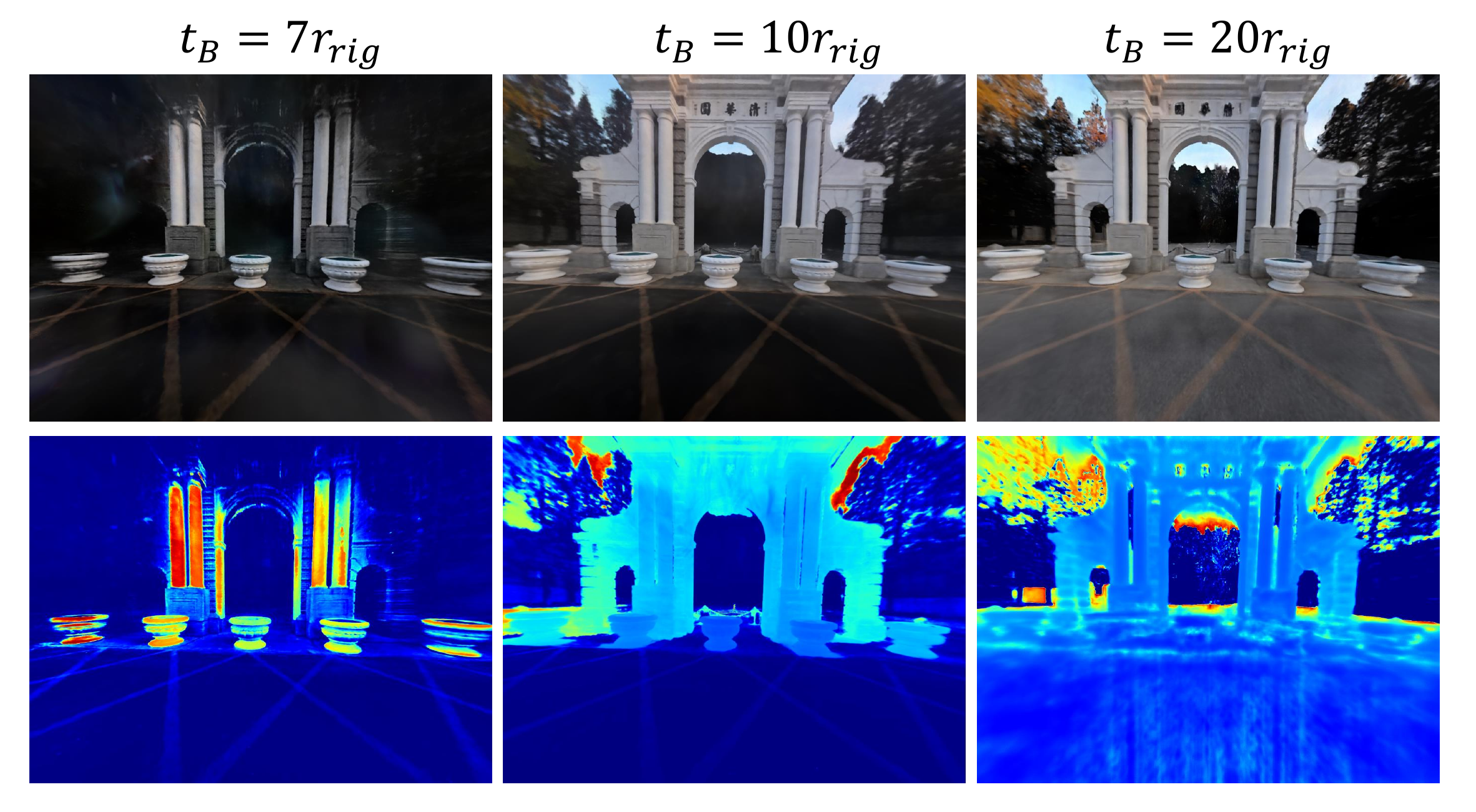}
\captionsetup{justification=centering}
\caption{Ablation on boundary assignment.}
\label{figure: ablation}
\end{figure}
\subsubsection{Effect of boundary assignment}
Boundary assignment depends on cases. Some complex scenes in the THUImmersive dataset are sensitive to the position of the boundary. Fig. \ref{figure: ablation} shows some different boundary assignments on the foreground and background separation on the THUImmersive dataset. $t_B$ indicates the boundary depth, and $r_{rig}$ indicates the radius of our camera rig. The first row shows the rendered color map of the foreground scene, and the second shows the foreground depth map. As shown in the first column, shallow boundary leads to much more floaters at vacant space. When the boundary is set to be excessively deep, as shown in the last column, it's hard to capture high-frequency details, leading to blurred outcomes. Therefore, in this case, we reckon $t_B=10r_{rig}$ to be the appropriate boundary depth.

\section{Conclusion}
ImmersiveNeRF proposes a novel foreground-background hybrid representation for efficient large-space immersive light field reconstruction, which mainly focuses on unbounded scenes captured by an inside-looking-out configuration.
We leverage spherical scene representation for foreground and background respectively. Specifically, we first separate the foreground and background into different radiance fields to efficiently learn the novel view synthesis in the large-scale scene. Based on this, we further propose an adaptive sampling strategy and a segmentation regularizer for more robust convergence. We also capture a novel immersive light field dataset, named THUImmersive, with the potential to achieve a much larger scale 6DoF immersive rendering effects to support related research.
We qualitatively and quantitatively compare our approach to prior methods aiming at large-scene rendering, showing that ImmersiveNeRF outperforms these methods.

\textbf{Future Work.}
During the experiment, we found out that the accuracy of camera poses affects the rendering quality. 
Also, the capture setting that cameras shot in an outward manner leads to different camera imaging configurations, such as different white balance and exposure, which have an impact on the robustness of reconstruction, especially for real-world scenes. 
Therefore, in future work, we aim to introduce the robust ImmersiveNeRF under inaccurate camera poses and inconsistent imaging configurations for future utilization in the practical scene and also push the rendering speed to real-time for immersive AR/VR applications.

\vfill
\bibliographystyle{IEEEtran}
\bibliography{aaai24}

\end{document}